%% file: tacl2021v1-template_FOR_ARXIV.tex
\documentclass[11pt,a4paper]{article}

\usepackage{times,latexsym}
\usepackage{url}
\usepackage[T1]{fontenc}

\usepackage[acceptedWithA]{tacl2021v1}

\usepackage{xspace,mfirstuc,tabulary}

\definecolor{Purple}{RGB}{255,10,140}

\newcommand{\ex}[1]{{\sf #1}}

\newif\iftaclinstructions
\taclinstructionsfalse 
\iftaclinstructions
\renewcommand{\confidential}{}
\renewcommand{\anonsubtext}{(No author info supplied here, for consistency with
TACL-submission anonymization requirements)}
\newcommand{\instr}
\fi

\iftaclpubformat 

\else

\fi

\usepackage{graphicx}
\usepackage{subcaption}
\usepackage{tikz}
\usepackage{booktabs}
\newcommand{\scalemate}[1]{\emph{#1}}
\newcommand{\scale}[2]{$\langle$\scalemate{#1}, \scalemate{#2}$\rangle$}
\newcommand{\blank}{\rule{1cm}{0.15mm}}
\newcommand{\blob}[1]{
    \vcenter{\hbox{\begin{tikzpicture}
    \node[rounded corners,fill=black!5] {\small\texttt{#1}};
    \end{tikzpicture}}}
}
\newcommand{\weak}{{\color{teal}\texttt{[WEAK]}}}
\newcommand{\strong}{{\color{red}\texttt{[STRONG]}}}
\newcommand{\context}{{\color{black!40}{\texttt{[CONTEXT]}}}}
\newcommand{\scalartemplate}{
    {\color{black!40}{\small\texttt{[CONTEXT]}}}~
    \underbrace{\blob{\weak\color{orange}{, but not} \strong\color{orange}{,}}}_{\text{\emph{scalar construction}}}~
    {\color{black!40}{\small\texttt{[CONTEXT]}}}
}
\newcommand{\weakcolor}[1]{{\color{teal}{#1}}}
\newcommand{\strongcolor}[1]{{\color{red}{#1}}}
\usepackage{inconsolata}
\usepackage{amsmath}
\usepackage{cleveref}
\usepackage{centernot}
\newcommand\blfootnote[1]{%
  \begingroup
  \renewcommand\thefootnote{}\footnote{#1}%
  \addtocounter{footnote}{-1}%
  \endgroup
}
\usepackage{bm}
\usepackage{multirow}
\usepackage{linguex}

\title{Expectations over Unspoken Alternatives Predict Pragmatic Inferences}

\author{
  Jennifer Hu$^1$, Roger Levy$^1$, Judith Degen$^2$,
  \and
  Sebastian Schuster$^3$
  \\
  \ \\
  $^1$Department of Brain and Cognitive Sciences, MIT, \texttt{\{jennhu, rplevy\}@mit.edu}
  \\
  $^2$Department of Linguistics, Stanford University, \texttt{jdegen@stanford.edu} \\
  $^3$Department of Language Science, Saarland University, \texttt{seschust@lst.uni-saarland.de}
}

\date{}

\begin{document}
\maketitle
\begin{abstract}
  Scalar inferences (SI) are a signature example of how humans interpret language based on unspoken alternatives. While empirical studies have demonstrated that human SI rates are highly variable -- both within instances of a single scale, and across different scales -- there have been few proposals that quantitatively explain both cross- and within-scale variation. Furthermore, while it is generally assumed that SIs arise through reasoning about unspoken alternatives, it remains debated whether humans reason about alternatives as linguistic forms, or at the level of concepts. Here, we test a shared mechanism explaining SI rates within and across scales: context-driven expectations about the unspoken alternatives. Using neural language models to approximate human predictive distributions, we find that SI rates are captured by the expectedness of the strong scalemate as an alternative. Crucially, however, expectedness robustly predicts cross-scale variation only under a meaning-based view of alternatives. Our results suggest that pragmatic inferences arise from context-driven expectations over alternatives, and these expectations operate at the level of concepts.\blfootnote{Code and data can be found at \url{https://github.com/jennhu/expectations-over-alternatives}.} 
\end{abstract}

\section{Introduction}
Much of the richness of linguistic meaning arises from what is left unsaid \citep[e.g.,][]{grice_logic_1975,sperber_relevance_1986,horn_natural_1989}. For example, if Alice says ``Some of the students passed the exam'', Bob can infer that Alice means \emph{not all} students passed the exam, even though Alice's utterance would still be logically true if all students had passed. One explanation of this inference is that Bob reasons about the unspoken \textbf{alternatives} that were available to the speaker. Under the assumptions that (1) speakers generally try to be informative, (2) Alice has full knowledge of the situation, and (3) it would have been relevant and more informative for Alice to say ``All of the students passed the exam'', Alice's choice to say ``some'' suggests that she believes the sentence with ``all'' is false. This inference pattern is more generally known as \textbf{scalar inference} (SI), which arises from orderings between linguistic items (scales). 

\input{figure1.tex}
SI has often been treated as a categorical phenomenon: when a speaker utters a weaker (less informative) item on a scale, a listener rules out the meaning of stronger (more informative) items on that scale \citep[e.g.,][]{levinson_presumptive_2000}. However, empirical studies have demonstrated substantial variability in the rates at which humans draw SIs, both within instances of a single scale \citep{degen_investigating_2015,eiteljoerge_pieces_2018,li_predicting_2021} and across scales formed by different lexical items \citep[e.g.,][]{doran_non-unified_2009,beltrama_is_2013,van_tiel_scalar_2016,gotzner_scalar_2018,pankratz_role_2021,ronai_three_2022}. 
For example, consider the following instances of the scale \scale{some}{all}:

\begin{small}
\ex. 
    \a. I like some country music. \label{ex:some-country}
    \b. I like some, but not all, country music. \label{ex:sbna-country}

\end{small}

\begin{small}
\ex.
    \a. It would certainly help them to appreciate some of the things that we have here. \label{ex:some-appreciate}
    \b. It would certainly help them to appreciate some, but not all, of the things that we have here. \label{ex:sbna-appreciate}

\end{small}

\noindent \citet{degen_investigating_2015} finds that humans are highly likely to consider \ref{ex:some-country} as conveying a similar meaning as \ref{ex:sbna-country}, but unlikely to consider \ref{ex:some-appreciate} as conveying a similar meaning as \ref{ex:sbna-appreciate} (\Cref{fig:example-within-scale}). Similarly, consider the following instances of the scales \scale{possible}{certain} and \scale{ugly}{hideous}, which both consist of adjectives ordered by entailment: 

\begin{small}
\ex. 
    \a. Success is possible. \label{ex:possible}
    \b. Success is not certain. \label{ex:certain}

\end{small}

\begin{small}
\ex. 
    \a. The painting is ugly. \label{ex:ugly}
    \b. The painting is not hideous. \label{ex:hideous}
    
\end{small}

\noindent \citet{van_tiel_scalar_2016} find that humans are highly likely to conclude that \ref{ex:possible} implies \ref{ex:certain}, but unlikely to conclude that \ref{ex:ugly} implies \ref{ex:hideous} (\Cref{fig:example-cross-scale}). 

While cross-scale and within-scale variation have typically been studied as distinct empirical phenomena, they both reflect gradedness in listener inferences based on alternatives and context. It therefore seems desirable to explain these empirical findings with a shared account, but there have been few proposals that quantitatively explain both within- and cross-scale variation.
For example, cross-scale variation can be explained by intrinsic properties of the scale \citep[e.g., whether the strong scalemate refers to an extreme endpoint;][]{van_tiel_scalar_2016}, but these factors cannot explain variation within instances of a single scale. On the other hand, many factors explaining within-scale variance are scale-specific \citep[e.g., the partitive ``of the'' for \scale{some}{all};][]{degen_investigating_2015} and may not generalize to new scales.

Here, we investigate a shared account of SI rates within and across scales. Since the alternatives are not explicitly produced (by definition), the listener has uncertainty over which alternatives the speaker could have used -- and therefore, which strong scalemates ought to be negated through SI. Building upon constraint-based accounts of human language processing  \citep{degen_processing_2015,degen_availability_2016}, we test the hypothesis that SIs depend on the availability of alternatives, which depend on context-driven expectations maintained by the listener. For example, if a speaker says ``The movie was good'', the listener might predict that \scalemate{amazing} is a more likely alternative than \scalemate{funny} to the weak term \scalemate{good}. An expectation-based view predicts that the listener would be thus be more likely to infer that the movie is not amazing (according to the speaker), and less likely to infer that the movie is not funny. However, while \citet{degen_processing_2015,degen_availability_2016} have argued that listeners maintain context-driven expectations over alternatives, these studies have primarily investigated a single scale (\scale{some}{all}) in small domains, arguing from qualitative patterns and in the absence of a formal theory.

\begin{table*}[ht]
    \centering
    \scriptsize
    \begin{tabular}{llrrrr}\toprule
        Dataset & Type of variation & \# participants & \# scales & \# contexts per scale & \# data points per item \\ \midrule
        \citet{degen_investigating_2015} & Within-scale & 243 & 1 & 1363 & $\sim$ 10 \\ \midrule
        \citet{ronai_three_2022} & Cross-scale & 40 & 57 & 1 & 40 \\
        \citet{pankratz_role_2021} & Cross-scale & 1970 & 50 & 1 & $\sim$ 40 \\
        \citet{gotzner_scalar_2018} & Cross-scale & 220 & 67 & 1 & 40 \\
        \citet{van_tiel_scalar_2016} & Cross-scale & 28 & 39 & 3 & 10 \\ \bottomrule
    \end{tabular}
    \caption{Details of human data used in our analyses. An item is a unique (scale, context) combination.}
    \label{tab:datasets}
\end{table*}

Furthermore, while it is generally assumed that SIs arise based on reasoning about unspoken alternatives, it remains debated whether humans reason about alternatives as linguistic structures \citep[e.g.,][]{katzir_structurally-defined_2007,fox_characterization_2011}, or at the level of concepts \citep[e.g.,][]{gazdar_pragmatics_1979,buccola_conceptual_2021}. Returning to the earlier example, if the weak scalemate is \scalemate{good}, listeners may reason about a concept like \textsc{VeryGood} instead of a specific linguistic expression like \scalemate{amazing}. In this sense, the listener's uncertainty about alternatives might arise from uncertainty about both the scale itself (\emph{Is the speaker implying the plot wasn't amazing, or that the jokes weren't funny?}), as well as the exact word forms under consideration by the speaker (\emph{Is the speaker implying the movie wasn't amazing, fantastic, or wonderful?}). Despite theoretical debates about the nature of alternatives, however, the role of concept-based alternatives in SI has not been tested in a systematic, quantitative way.

We provide a formalization of an expectation-based account of alternatives and test it on both string-based and concept-based views of alternatives. Instead of empirically estimating human expectations over alternatives \citep[cf.][]{ronai_three_2022}, we use neural language models as an approximation, which allows us to generate predictions for arbitrary sentences and contexts. We test the account's predictions on human SI rates within the \scale{some}{all} scale \citep{degen_investigating_2015}, and across 148 scales from four datasets \citep{van_tiel_scalar_2016,gotzner_scalar_2018,pankratz_role_2021,ronai_three_2022}. We find support for the expectation-based account, and also provide the first evidence that concept-based alternatives may be underlying a wide range of SIs. Our results suggest that pragmatic inferences may arise from context-driven expectations over unspoken alternatives, and these expectations operate at the level of concepts.

\section{Background}

\subsection{Within-scale variation} \label{sec:bg-within-scale}

Within-scale variation refers to the variation in SI rates across instances of a single scale, such as \scale{some}{all}. To explore SI variation within the scale \scale{some}{all}, we use the dataset collected by \citet{degen_investigating_2015}, which features 1363 naturalistic sentences containing a ``some''-NP from the Switchboard corpus of telephone dialogues \citep{godfrey_switchboard_1992} (\Cref{tab:datasets}). For each sentence, SI rates were measured using a sentence-similarity paradigm. On each trial, participants saw two sentence variants: the original sentence containing ``some'', and a minimally differing sentence where ``, but not all,'' was inserted directly after ``some''. Participants were asked, ``How similar is the statement with `some, but not all' to the statement with `some'?'' and indicated responses (similarity judgments) on a seven point Likert scale. If the speaker's originally intended meaning clearly includes an implicature, then making the implicature explicit by inserting ``, but not all,'' should not change the meaning of the sentence, so similarity judgments should be high. Thus, a higher similarity judgment indicates a stronger SI.

\citet{degen_investigating_2015} finds substantial variation in SI rates across contexts, challenging the idea that the ``some, but not all'' inference arises reliably without sensitivity to context \citep{horn_natural_1989,levinson_presumptive_2000}. She also reports several features that predict SI rates, such as whether ``some'' occurs with the partitive ``of the'', or whether the ``some''-NP is in subject position. However, these features may be highly specific to the \scale{some}{all} scale, and it is unclear whether a more general mechanism may also explain variation within or across other scales.

\subsection{Cross-scale variation (scalar diversity)} \label{sec:bg-cross-scale}

Cross-scale variation refers to the variation in SI rates across scales formed by different lexical items. To explore this, we use SI rates across 148 unique scales from four datasets, summarized in \Cref{tab:datasets}. Each scale involves a pair of English words (adjectives, adverbs, or verbs) of the form \scale{\weak}{\strong}, where \weak{} is less informative than \strong{} (e.g., \scale{intelligent}{brilliant}).\footnote{We excluded scales where one of the items was formed by a multi-word expression (e.g., \scale{may}{have to}).} For each dataset, SI rates were measured through a binary choice task. Participants saw a character make a short, unembedded statement consisting of a simple noun phrase subject and a predicate with a weak scalar item (e.g., ``John says: This student is intelligent.''). Their task was to indicate (\emph{Yes} or \emph{No}) whether they would conclude that the speaker believes the negation of a strong scalar item (e.g., ``Would you conclude from this that, according to John, she is not brilliant?''). The SI rate for a scale is the proportion of \emph{Yes} responses. 

This method has revealed large variation in SI rates, ranging from 4\% (\scale{ugly}{hideous}) to 100\% (\scale{sometimes}{always}) \citep{van_tiel_scalar_2016}. \citet{van_tiel_scalar_2016} test two classes of factors that might predict SI rates: the availability of the strong scalemate given the weak scalemate, and the degree to which scalemates can be distinguished from each other. They find SI rates are predicted by measures of scalemate distinctness (e.g., whether the strong scalemate forms a fixed endpoint on the scale), but not by availability \citep[but see][]{westera_closer_2020,ronai_three_2022}. Other studies have proposed additional scale-intrinsic factors \citep[e.g.,][]{gotzner_scalar_2018,sun_link_2018,pankratz_role_2021}. However, structural properties of a scale cannot explain variablity in SI rates \emph{within} a scale, as these properties do not change across contexts. 

While others have proposed context-dependent factors -- which could, in principle, explain both cross- and within-scale variation -- these factors often lack explanatory power in practice. For example, \citet{ronai_exploring_2021} find that the prominence of the Question Under Discussion \citep{roberts_information_2012} is correlated with SI rates, but only for unbounded scales (i.e., scales where neither scalemate has a fixed, extreme meaning).

\section{An expectation-based account of SI} \label{sec:predictors}

Theoretically, it is the set of alternative utterances -- utterances that the speaker could have used, but didn't -- that drive scalar implicature, and in principle every possible utterance in a language might be an alternative to every other. However, at an algorithmic level \citep{marr_vision_1982}, it would be intractable for listeners to perform inference over this entire set. Furthermore, the signature pattern of SI would not arise without restrictions on the alternatives: otherwise, ``\weak{}, but not \strong{}'' and ``\strong{}'' would both be alternatives to ``\weak{}'', leading to contradictory inferences without a mechanism for breaking symmetry \citep{kroch_lexical_1972,katzir_structurally-defined_2007,breheny_symmetry_2018}. 

To solve this symmetry problem, some approaches restrict alternatives based on structural complexity through grammar-internal mechanisms \citep[e.g.,][]{katzir_structurally-defined_2007,fox_characterization_2011}. However, these theories do not capture the uncertainty that listeners maintain, and are difficult to test quantitatively. Here, we test the view that listeners form probabilistic expectations over alternatives, given information from their interaction with the speaker. In the remainder of this section, we first discuss the conceptual predictions of an expectation-based account of SI, and then describe how we operationalize these predictions using neural language models. 

Suppose that a listener hears a sentence with a weak scalar term \weak{} (e.g., ``This student is intelligent''). To rule out the meaning of a particular strong scalemate \strong{} (e.g., the student is not \scalemate{brilliant}), the listener must have reason to believe that the speaker would have said \strong{} if they had intended to convey the strong meaning. However, since the alternatives are not explicitly produced, the listener has some degree of uncertainty over what alternatives were considered by the speaker. If it is likely that the speaker would have said \strong{} to convey the strong meaning, then their choice to say \weak{} suggests that they did not have grounds to say \strong{} -- and thus, an SI should be more likely to arise.

The key question, then, is how listeners estimate which alternatives are likely to be considered by the speaker. An expectation-based account proposes that listeners integrate contextual and grammatical cues to maintain probabilistic expectations over these alternatives. A scalemate that is more probable (given these cues) should be more likely to enter the scalar inference computation. Thus, this account predicts that the more expected the strong scalemate is as an alternative to the weak scalemate, the higher SI rates should be.

\subsection{String-based view of alternatives} \label{sec:predictors-surprisal}

When an alternative is likely to be a strong scalemate, listeners should be more likely to rule out its meaning, resulting in higher SI rates. Conditioned on the context and the speaker's choice to use \weak{}, the listener must estimate the probability of \weak{} and \strong{} being contrasted in a scalar relationship. Since it is difficult to directly estimate this probability,  we construct a sentence frame where the probability of \strong{} -- at the level of forms -- approximates the probability of \strong{} being in a scalar relationship with a weak scalemate \weak{}. This approach allows us to re-frame the problem of estimating listeners' expectations over strong scalemates into a word prediction problem.

To do this, we use the scalar construction ``\emph{X, but not Y}'', which in many cases suggests that \emph{Y} is a strong scalemate to \emph{X} \citep{hearst_automatic_1992,de_melo_good_2013,van_miltenburg_detecting_2015,pankratz_role_2021}. For a given utterance \context{} \weak{} \context{} and hypothesized scale \scale{\weak{}}{\strong{}}, we form a sentence that explicitly states the SI: 
\begin{equation}
    \scalartemplate \label{eq:scalar-template}
\end{equation}

To test how expected \strong{} is as an alternative to \weak{}, we need to estimate how likely a human would predict \strong{} to appear in the \strong{} position in (\ref{eq:scalar-template}).\footnote{Another approach would be to measure the expectedness of \strong{} in the template \context{} \strong{} \context{} -- that is, by replacing \weak{} with \strong{} in the speaker's original utterance. This template would instantiate the theory that listeners determine alternatives based on the context. In contrast, the template we use in (\ref{eq:scalar-template}) instantiates the theory that listeners form expectations over alternatives based on the context as well as the speaker's usage of \weak. We return to this topic in \Cref{sec:discussion-template}. \label{fn:templates}}
Instead of attempting to directly measure these predictions \citep[cf.][see (\ref{eq:rx-cloze})]{ronai_three_2022}, we approximate this with neural language models. We measure how unexpected \strong{} is by computing its surprisal (negative log probability) under a language model, conditioned on the rest of the sentence. Since surprisal measures \emph{un}expectedness, we predict a negative relationship between SI rate and the surprisal of the strong scalemate. 

This predictor is closely related to the notion of an SI's ``relevance'' \citep{pankratz_role_2021}. Under usage-based theories of language \citep[e.g.,][]{tomasello_constructing_2003,bybee_usage-based_2015}, if a weak scalar term is encountered frequently in a scalar relationship with a particular strong term, then the scalar relationship between these items will be enforced. Thus, \citet{pankratz_role_2021} measure the relevance of an SI by counting corpus frequencies of the scalemates in the string ``\weak{}, but not \strong{}.'' This is conceptually aligned with our setup, where we might expect higher corpus frequencies to correspond to lower surprisal under a language model. However, our predictor differs from \citeauthor{pankratz_role_2021}'s in an important way: they aim to measure the ``general relevance'' of an SI, which they define as ``relevance even in the absence of a situated context.'' It is unclear how general relevance can explain variation in SI rates within instances of a scale. By using context-conditioned probabilities from a language model, our predictor could account for both the general frequency of ``\weak{}, but not \strong{}'' as well as expectations driven by the context in which the scale occurs. 

\subsection{Concept-based view of alternatives} 
\label{sec:predictors-concept-based-suprisal}
The method described above implicitly treats linguistic forms as the alternatives driving scalar inferences. However, recent proposals have advanced the view that alternatives are not linguistic objects, but instead operate at the level of more general reasoning preferences \citep{buccola_conceptual_2021}. On this view, alternatives are constructed by replacing primitives of the concept expressed by the speaker with primitives of equal or less complexity. 

Here, we test a generalization of this concept-based view of alternatives. Suppose, for example, a speaker uses the weak scalar term \scalemate{big}. On a concept-based view, the listener may infer that the speaker is contrasting \scalemate{big} with a concept like \textsc{VeryBig} instead of a particular linguistic expression like \scalemate{enormous}. 
However, in the experiments mentioned in \Cref{sec:bg-cross-scale}, the SI process likely needs to be grounded in linguistic forms before the listener makes a judgment about a particular strong scalemate (in string form). One hypothesis is that upon hearing an expression with a weak scalemate, a stronger conceptual alternative is activated, which in turn probabilistically activates all the strings that could reflect it. Returning to our earlier example, if the conceptual alternative is \textsc{VeryBig}, and \scalemate{huge}, \scalemate{massive}, and \scalemate{enormous} are string-based realizations of that alternative, they may be assigned a high likelihood. When asked about a specific string-form alternative (e.g., ``The elephant is big. Would you conclude that it is not enormous?''), humans may endorse the SI if the probability of conceptually similar linguistic alternatives is sufficiently high, even if the probability of the tested alternative (here, \scalemate{enormous}) is low. 

If SIs involve reasoning about conceptual alternatives, then surprisal values estimated from assumed string-form alternatives may be poor estimates of the true relevant surprisal, as a single concept could be expressed with multiple forms. Therefore, in addition to assessing whether expectedness of specific linguistic forms predicts SI rates (\Cref{sec:predictors-surprisal}), we also test a second predictor which approximates the expectedness of conceptual alternatives. To do this, we need a set of alternatives $\mathcal{A}$ that could serve as potential linguistic scalemates. As described in more detail in \Cref{sec:within-scale-alternatives,sec:cross-scale-alternatives}, we obtain $\mathcal{A}$ by taking a fixed set of words with the same part of speech as the weak scalemate, inspired by grammatical theories of alternatives \citep[e.g.,][]{rooth_association_1985,katzir_structurally-defined_2007}.\footnote{We adopt a liberal view of alternatives to avoid undergeneration. However, an important open question is how alternatives are determined, which we leave for future work.} 

Using this alternative set $\mathcal{A}$, we compute the weighted average surprisal of $\mathcal{A}$ using weights determined by the conceptual similarity between each alternative and the tested strong scalemate. We use GloVe embeddings \citep{pennington_glove_2014} as an approximation for conceptual representations of scalar items, and cosine similarity between GloVe vectors to approximate conceptual similarity.

For each scale \scale{\weak}{\strong}, we obtain weights by computing the cosine similarity between the GloVe embeddings for \strong{} ($v_{\strong}$) and each potential alternative $a$ ($v_a$) in the alternative set $\mathcal{A}$. We compute the weighted average probability over $\mathcal{A}$ using these weights, and then take the negative log to obtain the weighted average surprisal:
\begin{equation}
    -\log\left(\frac{\sum_{a\in\mathcal{A}} P(a) \cdot \text{cossim}(v_{\strong}, v_a)}{\sum_{a\in\mathcal{A}} \text{cossim}(v_{\strong}, v_a)}\right) \label{eq:weighted-surprisal}
\end{equation}
If there are many conceptually similar alternatives with low surprisal, then the weighted average surprisal will be low, even if the surprisal of the tested scalemate is high. Therefore, weighted average suprisal forms a proxy for concept-based surprisal, which we compare to string-based suprisal.

\section{Predicting variation within \scale{some}{all}} \label{sec:within-scale}

\subsection{Human data}

To investigate variation within the scale \scale{some}{all}, we use human SI strength ratings collected by \citet{degen_investigating_2015}. These ratings were measured by asking participants to rate the similarity (1-7) between a sentence with ``some'' and a minimally differing sentence with ``some, but not all''. See \Cref{sec:bg-within-scale} for details.

\subsection{Model}

Following the experiment conducted by \citet{degen_investigating_2015}, we construct scalar templates by inserting ``, but not all,'' after the occurrence of ``some'' in each sentence from the dataset. Since this scalar construction (``some, but not all,'') often occurs in the middle of the sentence, we use the bidirectional language model BERT \citep{devlin_bert_2019} to measure model expectations at the position of the strong scalemate. Concretely, we replace ``all'' with the \texttt{[MASK]} token and measure BERT's probability distribution at that token. All models in our study are accessed via the Huggingface \texttt{transformers} library \citep{wolf_transformers_2020}.

\subsection{Candidate alternatives} \label{sec:within-scale-alternatives}

For our string-based surprisal predictor (\Cref{sec:predictors-surprisal}), we are only concerned with the surprisal of the alternative \scalemate{all} in the \strong{} position in (\ref{eq:scalar-template}). However, to compute our concept-based surprisal predictor (\Cref{sec:predictors-concept-based-suprisal}), we need a set of candidate alternatives that could potentially serve as the strong scalemates implied by the speaker. Since the alternatives to \scalemate{some} are highly constrained by the grammar, we manually constructed a set of English quantifiers that can be used in contrast to \scalemate{some}: \scalemate{each}, \scalemate{every}, \scalemate{few}, \scalemate{half}, \scalemate{much}, \scalemate{many}, \scalemate{most}, and \scalemate{all}.  

\subsection{Results}

\begin{figure}[t]
    \centering
    \subfloat[\label{fig:within-scale-surprisal}]{\includegraphics[width=0.5\linewidth]{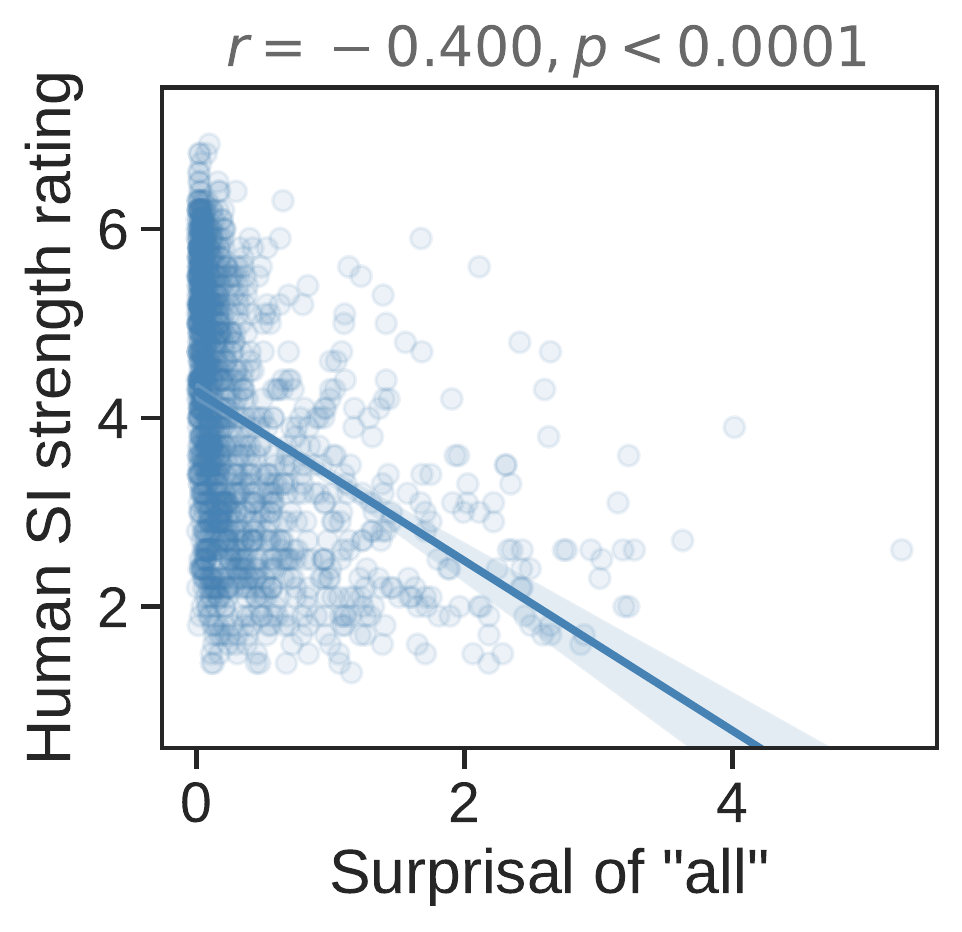}}~
    \subfloat[\label{fig:within-scale-concept-based-surprisal}]{\includegraphics[width=0.5\linewidth]{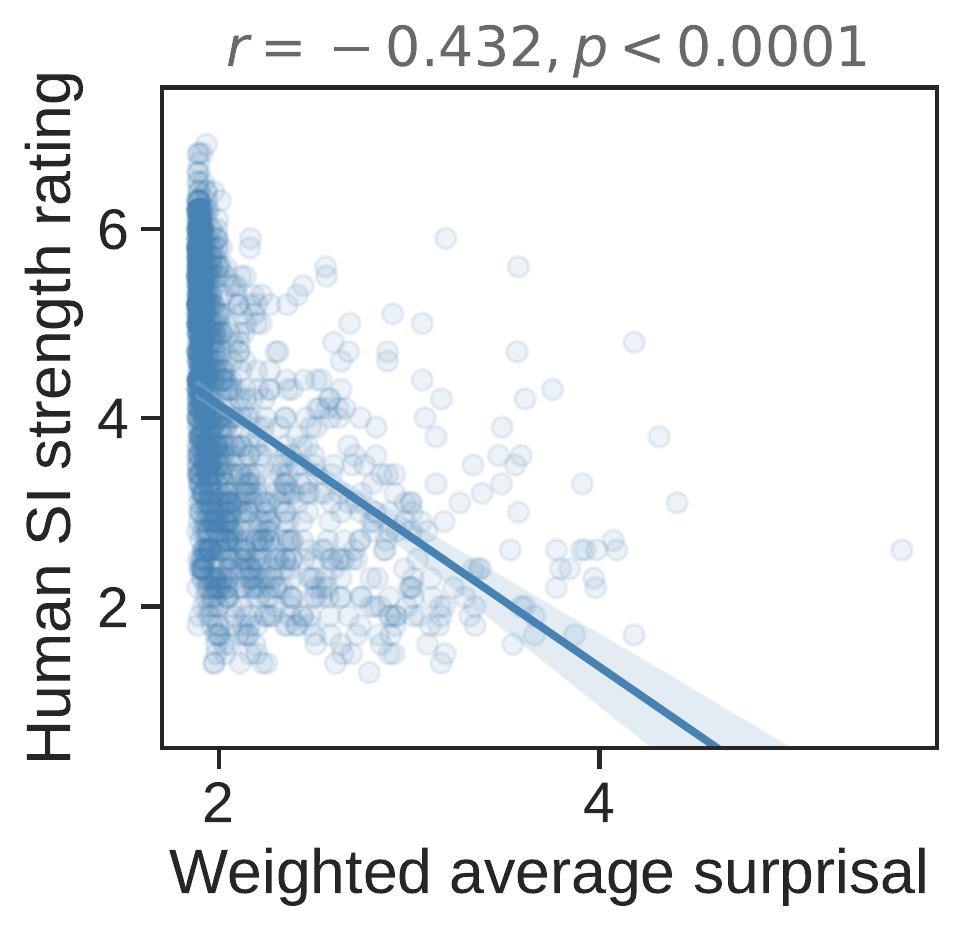}}
    \caption{Relationship between human SI strength ratings within \scale{some}{all} scale \citep{degen_investigating_2015} and BERT-derived predictors: (a) surprisal of scalemate \scalemate{all} in the scalar construction, and (b) weighted average surprisal over the full set of candidate alternatives (\Cref{sec:within-scale-alternatives}). 
    Each point represents a sentence. Shaded region denotes 95\% CI.
    }
    \label{fig:within-scale}
\end{figure}

\Cref{fig:within-scale} shows the relationship between our predictors and human SI ratings for \citeauthor{degen_investigating_2015}'s (\citeyear{degen_investigating_2015}) dataset of variation within \scale{some}{all}. We find that both string-based and concept-based surprisal are indeed negatively correlated with human similarity judgments (string-based: \Cref{fig:within-scale-surprisal}, Pearson $\rho=-0.400, p<0.0001$; concept-based: \Cref{fig:within-scale-concept-based-surprisal}, $\rho=-0.432, p<0.0001$).\footnote{We note that the relationship between surprisal and SI ratings appears highly non-linear in \Cref{fig:within-scale-surprisal}. We expect this is due to the fact that the scalemate \emph{all} is highly expected in most contexts, so the surprisal values of \emph{all} are concentrated near zero. There is a stronger linear relationship between SI ratings and raw probabilities ($\rho=0.482, p<0.0001$).}  

We additionally conducted a multivariate analysis including our two new predictors (string- and concept-based surprisal) among the predictors investigated in \citeauthor{degen_investigating_2015}'s original study. We centered and transformed all variables according to \citeauthor{degen_investigating_2015}'s original analyses. The results are summarized in \Cref{tab:within-scale}. We find that the original predictors remain statistically significant, and that concept-based surprisal (but not string-based surprisal) is a significant predictor in the full model. This suggests that listeners draw stronger scalar inferences when \scalemate{all} -- or a conceptually similar alternative -- is more expected in a given context.

\begin{table}[t]
    \centering
    \scriptsize
    \begin{tabular}{lrr} \toprule
        Predictor & $\beta$ & $p$ \\ \midrule
        \textsc{\citet{degen_investigating_2015} Predictors} & & \\
        Partitive & 0.658 & $<$ 0.0001 \\
        Strength & $-$0.470 & $<$ 0.0001 \\
        Mention & 0.287 & $<$ 0.0001 \\
        Subjecthood & 0.495 & $<$ 0.0001 \\
        Modification & 0.157 & $<$ 0.01 \\
        Log sentence length & 0.189 & $<$ 0.0001\\ \midrule
        \textsc{Our Predictors} & & \\
        String-based surprisal & \textbf{0.008} & \textbf{0.960} \\ 
        Concept-based surprisal & $\mathbf{-}$\textbf{0.782} & $\mathbf{<}$ \textbf{0.001} \\ \bottomrule
    \end{tabular}
    \caption{Summary of full regression model, including original predictors from \citet{degen_investigating_2015} (see the original study for a detailed description of each of the predictors).}
    \label{tab:within-scale}
\end{table}

\newcommand{\np}{{\color{black!60}\texttt{[NP]}}}
\newcommand{\adj}{{\color{black!60}\texttt{[ADJ]}}}
\begin{table*}[t]
    \centering
    \scriptsize
    \begin{tabular}{lllll} \toprule
        POS & \# unique & Form of original sentence & Form of scalar construction & Example \\ \midrule
        Adj & 120 & \np{} is \weak{} & \np{} is \weak{}, but not \strong{} & The elephant is \weakcolor{big}, but not \strongcolor{enormous} \\
        Adv & 12 & \np{} is \weak{} \adj{} & \np{} is \weak{} \adj{}, but not \strong{} & The director is \weakcolor{sometimes} late, but not \strongcolor{always} \\
        Verb & 16 & \np{} \weak{}-ed & \np{} \weak{}-ed, but did not \strong{} & The runner \weakcolor{started}, but did not \strongcolor{finish} \\ \bottomrule
    \end{tabular}
    \caption{Scalar construction templates for different parts of speech (for cross-scale variation).}
    \label{tab:cxs}
\end{table*}

\section{Predicting variation across scales} \label{sec:cross-scale}

\subsection{Human data}

To investigate variation across scales, we use human SI rates collected by four studies \citep{ronai_three_2022,pankratz_role_2021,gotzner_scalar_2018,van_tiel_scalar_2016}. SI rates were measured by showing participants a sentence with the weak scalemate (e.g., ``The student is intelligent''), and asking whether they would endorse the negation of the strong scalemate (e.g., ``The student is not brilliant''). See \Cref{sec:bg-cross-scale} for details.

\subsection{Model}

We construct scalar templates following the pattern summarized in \Cref{tab:cxs}. Since in each case the strong scalemate is the final word in the sentence,\footnote{For a small number of verbal scales, the strong scalemate is followed with the pronoun ``it'' to make the sentence grammatical. We don't expect this to matter for our purposes.} we use an autoregressive language model to measure expectations over potential scalemates in the \strong{} position. We use the base GPT-2 model \citep{radford_language_2019} via Huggingface and obtain model surprisals through the SyntaxGym command-line interface \citep{gauthier-etal-2020-syntaxgym}.

\subsection{Candidate alternatives}\label{sec:cross-scale-alternatives}

Recall from \Cref{sec:predictors-concept-based-suprisal} that we need a set of potential linguistic alternatives to compute the weighted average surprisal. We take this set of alternatives to be a set of words with the same part of speech (POS) as the weak scalemate and obtain these candidate alternative sets by extracting lists of English adjectives, adverbs, and verbs from WordNet \citep{miller_wordnet_1995}. We then used NLTK \citep{loper_nltk_2002} to find the words satisfying finer-grained POS tags (\texttt{JJ} for adjectives, \texttt{RB} for adverbs, and \texttt{VB} for verbs), and sorted each POS set according to word frequencies from the OpenSubtitles corpus \citep{lison_opensubtitles2016_2016}.\footnote{\url{https://github.com/hermitdave/FrequencyWords}}\textsuperscript{,}\footnote{\url{http://www.opensubtitles.org}} We excluded words in the POS sets that were not in the frequency corpus, resulting in 3204 adjectives, 1953 adverbs, and 226 verbs. We restricted each POS set to its 1000 highest-frequency words, and performed some manual exclusions (e.g., removing ``do'' and ``be'' from the verb set, which are unlikely to form scales with any of the tested items and follow different syntactic rules). This finally resulted in our three alternative sets: 1000 adjectives, 960 adverbs, and 224 verbs.\footnote{Most words in the alternative sets occur with low frequency, but we chose to be liberal when including alternatives to ensure broad coverage over potential scalemates.}

\begin{figure*}[ht]
    \centering
    \subfloat[\label{fig:cross-scale-surprisal}]{\includegraphics[width=0.8\linewidth]{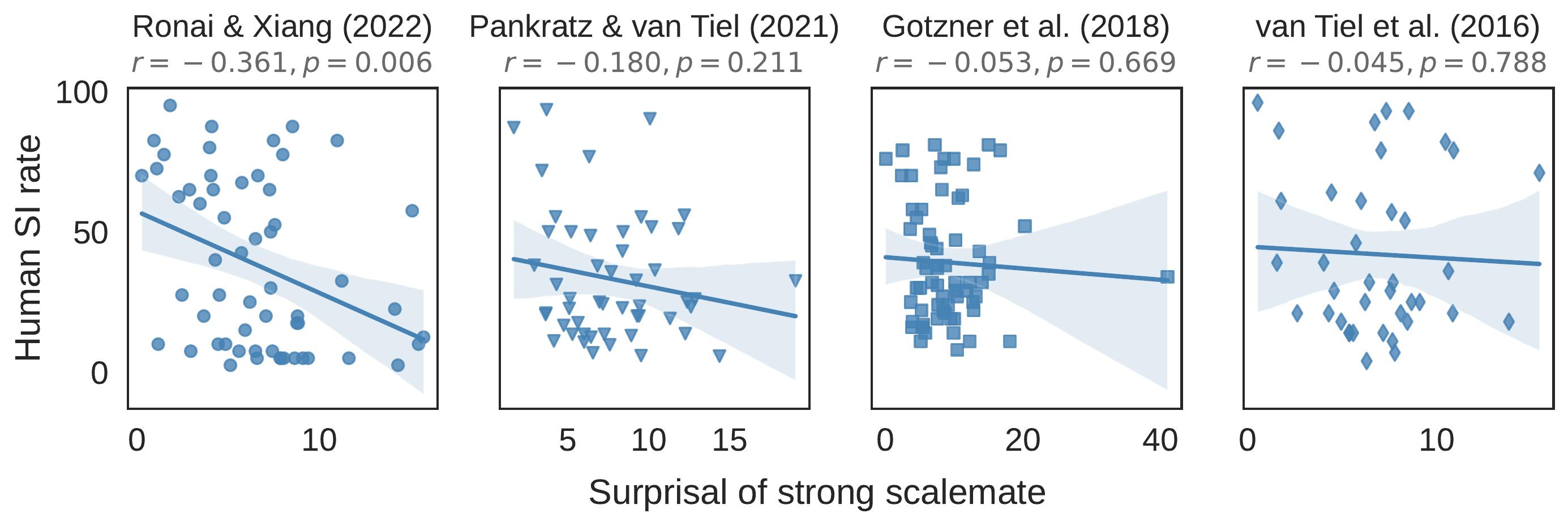}}\\
    \subfloat[\label{fig:cross-scale-weighted-avg-surprisal}]{\includegraphics[width=0.8\linewidth]{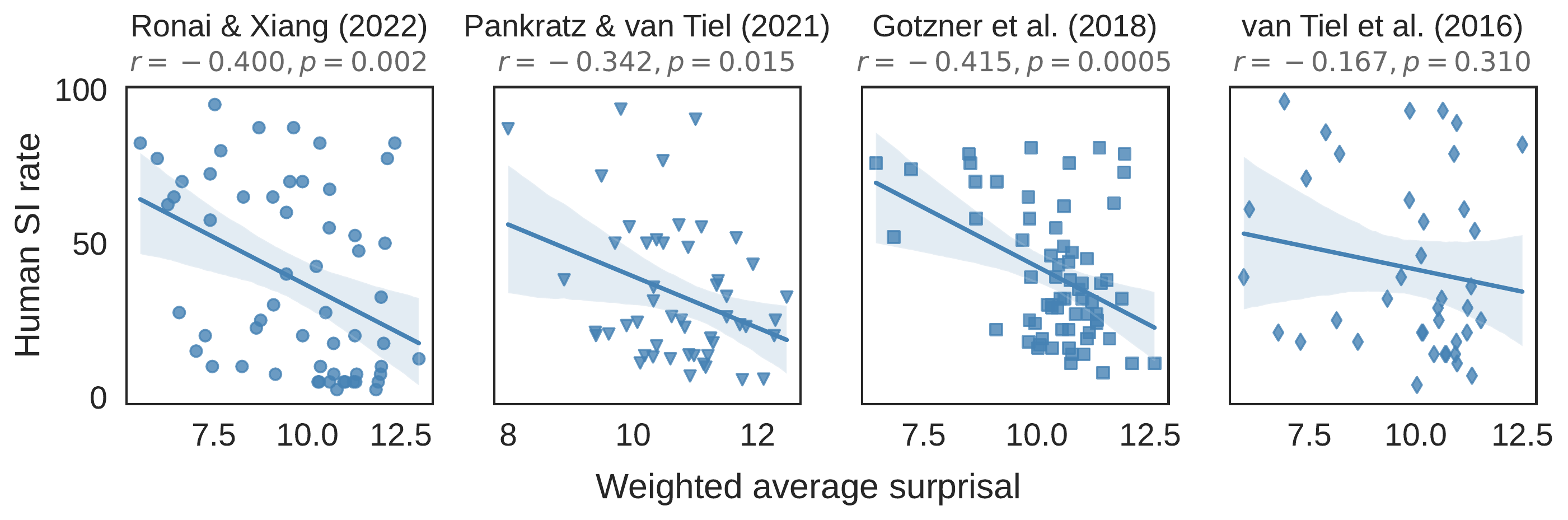}}
    \caption{Relationship between human SI rates and GPT-2-derived predictors across scales, for four datasets. Each point represents a single scale. Shaded region denotes 95\% CI. (a) SI rate vs.~surprisal of strong scalemate in the scalar construction. (b) SI rate vs.~weighted average surprisal over the full set of candidate alternatives (\Cref{sec:cross-scale-alternatives}).
    }
    \label{fig:cross-scale}
\end{figure*}

\begin{figure}[ht]
    \centering
    \includegraphics[width=0.55\linewidth]{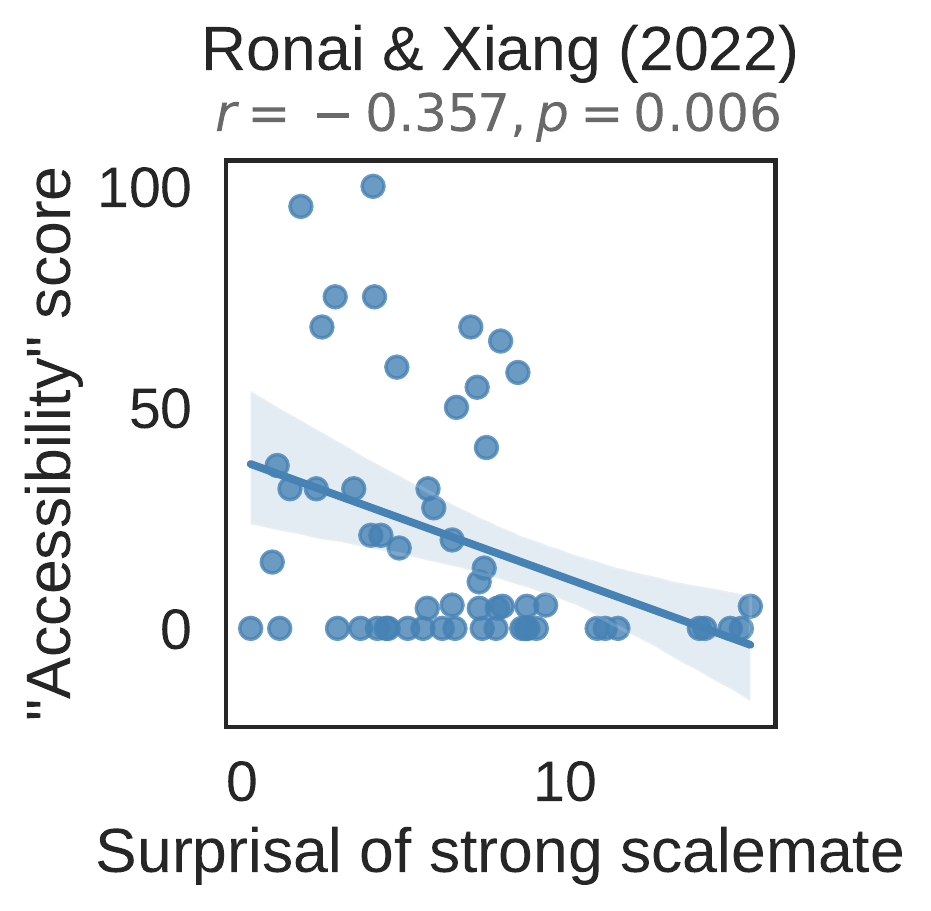}
    \caption{GPT-2-derived surprisal of strong scalemate vs.~accessibility rating of strong scalemates \citep{ronai_three_2022}.}
    \label{fig:surprisal-accessibility-rx22}
\end{figure}

\subsection{Results} \label{sec:cross-scale-main-results}

\subsubsection{String-based analyses} \label{sec:cross-scale-main-results-string}

\Cref{fig:cross-scale-surprisal} shows our results for cross-scale variation, under a string-based view of alternatives. We find that surprisal is a significant predictor only for \citeauthor{ronai_three_2022}'s dataset (Pearson $\rho=-0.361, p=0.006$).\footnote{We repeated this analysis after removing an outlier from \citeauthor{gotzner_scalar_2018}'s dataset, and again found a lack of relationship between SI rate and surprisal ($\rho=-0.0452, p=0.719$).} 

\begin{table*}[t]
   \centering
   \scriptsize
   \begin{tabular}{l|lrr|rr} \toprule
        \multirow{2}{*}{Dataset} & \multicolumn{3}{c}{Full model} & \multicolumn{2}{|c}{ANOVA} \\
        & Predictor & $\beta$ & $p$ & $F$ & $p$ \\ \midrule
       \multirow{2}{*}{\citet{ronai_three_2022}} & String-based surprisal & $-$1.538 & 0.215 & \multirow{2}{*}{3.247} & \multirow{2}{*}{0.012} \\ 
        & Concept-based surprisal & $-$4.503 & 0.065 & & \\ \midrule 
       \multirow{2}{*}{\citet{pankratz_role_2021}} & String-based surprisal & 0.460 & 0.694 & \multirow{2}{*}{3.198} & \multirow{2}{*}{0.050} \\ 
        & Concept-based surprisal & $-$9.491 & 0.036 & & \\ \midrule
       \multirow{2}{*}{\citet{gotzner_scalar_2018}} & String-based surprisal & 0.384 & 0.545 & \multirow{2}{*}{2.751} & \multirow{2}{*}{0.019}  \\ 
        & Concept-based surprisal & $-$8.010 & 0.0005 & & \\ \midrule
       \multirow{2}{*}{\citet{van_tiel_scalar_2016}} & String-based surprisal & 0.293 & 0.858 & \multirow{2}{*}{1.016} & \multirow{2}{*}{0.422}  \\ 
        & Concept-based surprisal & $-$3.340 & 0.291 & & \\\bottomrule
   \end{tabular}
   \caption{Summary of full regression model (middle columns) and ANOVA comparing full model against intercept-only model (right columns) for each cross-scale variation dataset.} \label{tab:cross-scale-mv}
\end{table*}

\paragraph{Model surprisal vs.~human completions.} For the dataset where we do find a relationship between surprisal and SI rates, we ask whether model surprisals are correlated with human-derived measurements of how ``accessible'' the strong scalemate is. If model surprisals and human accessibility scores are strongly linked, this would suggest that models and humans are aligned at the level of predictive distributions over alternatives, validating our approach of using language models to approximate human predictions.

To this end, we use data from \citeauthor{ronai_three_2022}'s Experiment 2, which measured the accessibility of scalemates through a Cloze task. Humans were presented with a short dialogue featuring a sentence with the weak scalemate, as in (\ref{eq:rx-cloze}), and then asked to generate a completion of the dialogue in the blank. The ``accessibility'' of the strong scalemate is taken to be the frequency with which it is generated in this paradigm.
\begin{align}
    \small\text{Sue:}\;\;\; &\small\text{The movie is good.} \label{eq:rx-cloze} \\ 
    \small\text{Mary:}\;\;\; &\small\text{So you mean it's not \blank.} \nonumber
\end{align}
We find that model surprisals are negatively correlated with accessibility scores (\Cref{fig:surprisal-accessibility-rx22}; $\rho=-0.357, p=0.006$), suggesting that our method of estimating expectations over alternatives using artificial language models aligns with direct measurements in humans.

\subsubsection{Concept-based analyses} \label{sec:cross-scale-conceptual-alts}

Turning to a conceptual view of alternatives, \Cref{fig:cross-scale-weighted-avg-surprisal} shows the relationship between human SI rates and weighted average surprisals (Equation \ref{eq:weighted-surprisal}). We find a significant negative correlation for all but one of the tested datasets (\citeauthor{ronai_three_2022}: $\rho=-0.400, p=0.002$; \citeauthor{pankratz_role_2021}: $\rho=-0.342, p=0.015$; \citeauthor{gotzner_scalar_2018}: $\rho=-0.415, p=0.0005$; \citeauthor{van_tiel_scalar_2016}: $\rho=-0.167, p=0.310$), demonstrating that similarity-weighted surprisal captures more variation than raw surprisal (cf.~\Cref{fig:cross-scale-surprisal}; \Cref{sec:cross-scale-main-results-string}). 

\input{rx_dists_figure.tex}

We additionally included both (centered) string-based and concept-based surprisal as predictors in a multivariate model, summarized in \Cref{tab:cross-scale-mv} (middle columns). As in the within-scale analysis, for three of the four datasets we find that concept-based surprisal is a stronger predictor than string-based surprisal. With that said, we find only a marginal effect of concept-based surprisal in \citeauthor{ronai_three_2022}'s data, and no effect of either predictor in \citeauthor{van_tiel_scalar_2016}'s data. However, for \citeauthor{ronai_three_2022}'s data, this does not mean that there is no value in either predictor -- rather, the predictors are too closely correlated to definitively favor one over the other. To demonstrate this, for each dataset we performed an analysis of variance (ANOVA) comparing the full model to a null intercept-only model (\Cref{tab:cross-scale-mv}, right columns). We find that for all datasets except that of \citeauthor{van_tiel_scalar_2016}, the model with both surprisal predictors explains significantly more variance than the null model. In sum, our results suggest that the expectedness of the strong scalemate can capture significant cross-scale SI variation, but these expectations may operate over groups of semantically similar linguistic forms instead of individual strings.

\paragraph{Qualitative analysis.} As a follow-up analysis, we identified cases where GPT-2 assigns low probability to the tested strong scalemate, but high probability to near synonyms. We analyzed the top 5 alternatives from the full alternative set (\Cref{sec:cross-scale-alternatives}) that were assigned highest probability as strong scalemates under GPT-2. \Cref{fig:rx-dists} shows three examples from \citeauthor{ronai_three_2022}'s dataset. The title of each subplot shows the scalar construction, with the weak scalemate highlighted in {\color{teal}{teal}} and the tested strong scalemate underlined in {\color{red}{\underline{red}}}. The y-axis shows the top 5 candidate scalemates, and the x-axis shows the probability assigned by the model. For the weak scalemate \emph{big} (left), GPT-2 assigns highest probability to the alternative \emph{huge}, which semantically conveys similar information to the empirically tested alternative \emph{enormous}. We see a similar pattern for weak scalemate \emph{largely} and alternatives \emph{completely} and \emph{totally} (middle), as well as for weak scalemate \emph{hard} and alternative \emph{impossible} (right). This is consistent with the hypothesis that surprisal of a specific string may not capture surprisal of the underlying concept.

Taken together, these analyses suggest that a concept-based view of alternatives is better aligned with human inferences than treating alternatives as specific linguistic forms. Testing additional ways of operationalizing concept-based alternatives is a promising direction for future work.

\section{Related work} \label{sec:related-work-nlp}

Prior work has evaluated the ability of computational models to capture scalar inferences. For example, the IMPPRES benchmark \citep{jeretic_are_2020} frames SI as a natural language inference problem: the weak scalar expression (e.g., ``Jo ate some of the cake'') is the premise, and the negated strong scalar expression (e.g., ``Joe didn't eat all of the cake'') is the hypothesis. Under this setup, an interpretation consistent with the strictly logical reading would assign a \emph{neutral} relationship between the premise and hypothesis, whereas a pragmatic reading would assign an \emph{entailment} relationship. Models are evaluated based on how often they assign the entailment label across items, which treats SIs as a homogeneous phenomenon and does not capture SI variation.

Another line of work has attempted to predict within-scale SI variation through a supervised approach \citep{schuster_harnessing_2020,li_predicting_2021}. This approach takes a sentence with a weak scalar item, and attempts to directly predict the human SI strength through a prediction head on top of a sentence encoder. This differs from our approach in that it requires training directly on the SI-rate-prediction task, whereas we probe the predictive distribution that emerges from language modeling with no task-specific representations. This allows us to compare model probability distributions to the expectations deployed by humans during pragmatic inferences, building upon a literature linking language models to predictive processing \citep[e.g.,][]{frank_insensitivity_2011,smith_effect_2013,wilcox_predictive_2020,merkx_human_2021}.

There have also been several studies extracting scalar orderings from corpora or language model representations. For example, \citet{de_marneffe_was_2010} use distributional information from a web corpus to ground the meanings of adjectives for an indirect question answering task. Similarly, \citet{shivade_corpus-based_2015} use scalar constructions like ``\emph{X, but not Y}'' to identify scales from a corpus of biomedical texts. Others have found that adjectival scale orderings can be derived from static word embeddings \citep{kim_deriving_2013} and contextualized word representations \citep{gari_soler_bert_2020,gari_soler_scalar_2021}. 

\section{Discussion}

We tested a shared mechanism explaining variation in SI rates across scales and within \scale{some}{all}, based on the hypothesis that humans maintain context-driven expectations about unspoken alternatives \citep{degen_processing_2015,degen_availability_2016}. We operationalized this in two ways using neural language models: the expectedness of a linguistic alternative as a scalemate (string-based surprisal), and the expectedness of a conceptual alternative (weighted average surprisal). We found that for both within-scale and cross-scale variation, expectedness captures human SI rates. Crucially, however, expectedness of the strong scalemate is a robust predictor of cross-scale variation only under a conceptual view of alternatives \citep{buccola_conceptual_2021}. Our results support the idea that the strength of pragmatic inferences depends on the availability of alternatives, which depends on in-context predictability.

One open question is the source of variability across the tested human behavioral datasets -- in particular, the lack of surprisal effect for \citeauthor{van_tiel_scalar_2016}’s data (\Cref{sec:cross-scale-main-results}). While we cannot be certain about why the results vary, we identified a few differences that might affect data quality across datasets (see \Cref{tab:datasets}). \citeauthor{van_tiel_scalar_2016}’s study has the smallest number of participants (28), smallest number of ratings per scale (10), and smallest number of scales (39). In addition, their experiments presented multiple sentence contexts per scale, whereas the other experiments only presented one sentence per scale. Other experimental factors, such as participant recruitment and exclusion criteria, may have also contributed to differences in data reliability.

\subsection{How do listeners restrict the alternatives?} \label{sec:discussion-template}

We now return to the issue raised in \Cref{fn:templates}: what information do listeners use to form expectations about alternatives? To illustrate potential hypotheses, consider the item ``The soup is warm/hot'' from \citeauthor{van_tiel_scalar_2016}'s experimental materials. In our framework described in \Cref{sec:predictors-surprisal}, $\context =$ ``The soup is'', $\weak = $ ``warm'', and $\strong =$ ``hot''. One hypothesis is that listeners form expectations over relevant scalar expressions given \context{} alone. On this view, expectations over strong scalemates could be measured by computing the probability of \strong{} in the template $\context \strong$; i.e., ``The soup is \strong''. In contrast, in this paper we test expectations of \strong{} in the template ``The soup is warm, but not \strong'', which instantiates an alternate theoretical position: that listeners use not only the context, but also \weak{} as information for forming expectations over alternatives. 

We adopt this view for several reasons. First, it could be the case that the context does not provide enough information for the listener to narrow down alternatives. Returning to the running example, ``The soup is'' could be followed by many continuations, some potentially relating to the taste or size of the soup in addition to its temperature. Taking the weak scalar term ``warm'' into account allows the listener to restrict the relevant alternatives to a smaller, more tractable set, which presents an algorithmic solution to the computationally challenging inference problem. However, the underinformativity of the context may be a problem unique to the simple stimuli used in the behavioral experiments. It is plausible that listeners could sufficiently restrict alternative sets given more naturalistic contexts, which likely provide more cues to the Question Under Discussion \citep{roberts_information_2012}. 

In addition, there could be cues from \weak{} that provide information about likely alternatives, independent of the context. For example, listeners might prefer strong scalemates that match \weak{} in register or formality, or in shared phonological features. This motivates why we chose template (\ref{eq:scalar-template}) to measure expectations over alternatives, instead of $\context \strong$. However, the extent to which listeners tune their predictions based on \weak{} above and beyond the context remains an open empirical question.

\subsection{From alternatives to inference}

Conceptually, computing an SI involves two steps: (1) determining the suitable alternatives, and (2) ruling out the meaning of alternatives to arrive at a strengthened interpretation of the weak scalar term. Our results primarily shed light on the first step, providing evidence that expectations play a role in determining alternatives, and that alternatives are likely based on meanings in addition to linguistic forms.

When considering the higher-level reasoning process, many factors beyond alternatives play a causal role in SI. One view is that humans use alternatives in a cooperative reasoning process, such as that formalized by the Rational Speech Act framework \citep[RSA;][]{frank_predicting_2012,goodman_pragmatic_2016}. In an RSA model, a pragmatic listener $L_1(m \mid u)$ uses a speaker's utterance $u$ to update their prior beliefs $P(m)$ over which meaning $m$ the speaker is trying to convey. The listener does this by computing the likelihood of a pragmatic speaker $S_1$ producing $u$ given each potential meaning. The pragmatic $S_1$ speaker corresponds to the utility $U$ of the utterance $u$ to convey $m$, relative to the utility of the alternative utterances in the set of alternatives $\mathcal{A}$:

\begin{align}
L_1(m \mid u) =  \frac{S_1(u \mid m)P(m)}{\sum_{m'}S_1(u \mid m')P(m')} \\
S_1(u \mid m) = \frac{U(u, m)}{\sum_{u' \in \mathcal{A}} U(u', m)}
\end{align}

Our findings appear compatible with RSA: listeners reason about a speaker that normalizes over alternatives. However, it remains an open question how variable expectations over alternatives should be operationalized in an RSA model. One option, as recently proposed by \newcite{zhang2023scalar}, is that the pragmatic speaker is conditioned on the alternative set $\mathcal{A}$. The pragmatic listener has beliefs over different sets of $\mathcal{A}$ and marginalizes over these beliefs when drawing an inference:

\begin{align}
L_1(m \mid u) = \sum_{\mathcal{A}} P(\mathcal{A})\frac{S_1(u \mid m, \mathcal{A})P(m)}{\sum_{m'}S_1(u \mid m', \mathcal{A})P(m')}
\end{align}

Another possibility is that the variable expectations are not inputs to the model, but instead fall out of reasoning about how likely speakers are to use the weaker versus stronger terms, given variable contextual priors over meanings and questions under discussion \citep[see, e.g.,][]{goodman_probabilistic_2015,qing_rational_2016}. We leave a detailed exploration of such a model to future work. 

\paragraph{The role of priors.} Pragmatic inferences are influenced by the prior probabilities of the world states compatible with the weak and strong meanings \cite{degen_wonky_2015,sikos_reevaluating_2021}. For example, consider the scale \scale{start}{finish}. If a human were asked ``The movie started at 2:30. Would you conclude that the movie did not finish at 2:30?'', they would likely answer \emph{Yes}. This \emph{Yes} response would count as an SI under the experimental paradigm, but does not reflect pragmatic reasoning over scalar alternatives: it is simply implausible for a movie to start and finish at the same time, given our knowledge of the world.\footnote{This example is due to \citet{lassiter_how_2022}.} 

These priors have an important connection to our analyses. As outlined in \Cref{sec:predictors-surprisal}, we approximate the expectedness of a strong scalemate by measuring the expectedness of its linguistic form. This approach can be seen as reflecting an implicit assumption that the more likely a certain meaning is, the more likely it is to be expressed linguistically. This is likely to be wrong in certain cases -- for example, if a certain meaning is so likely that it is obvious without being said, then speakers may avoid the effort of explicitly producing the linguistic expression (and thus, the linguistic expression would have low probability). This could potentially be the case for relatively common SIs. For example, a speaker might be able to get away with only saying \scalemate{some} and expecting a listener to recover the meaning \emph{some but not all}. 

With that said, we believe our estimation method may minimize this issue, as we measure expectations conditioned on an explicit scalar contrast with the weak scalemate (i.e., ``\weak, but not''). Thus, our approach can be seen as approximating listeners' expectations about upcoming linguistic material, given that the speaker has \emph{already chosen} to produce a scalar contrast. Nevertheless, a complete account of scalar inferences will need to account for the influence of the prior probabilities over world states, which may explain some of the variance not captured by our expectedness predictors.

\subsection{Implications for NLP}

While the main role of language models in our analyses was to systematically test a cognitive theory, we believe this work also has implications for NLP evaluation. A growing body of work uses controlled assessments to evaluate the linguistic knowledge of NLP models. Many studies test whether models exhibit a categorical pattern of behavior that reflects a particular linguistic generalization. For example, in syntactic evaluations, a model is successful if it satisfies certain inequality relationships between grammatical and ungrammatical sentences \citep[e.g.,][]{linzen_assessing_2016,futrell_neural_2019,hu_systematic_2020}. SI (and other types of implicatures) have largely been treated the same way (see \Cref{sec:related-work-nlp}). 

In contrast, we do not evaluate whether language models exhibit a categorical pattern of behavior (\emph{``Do models interpret SIs pragmatically?''}). Instead, based on the empirical evidence for scalar variation, we test whether models capture systematic variability in human inferences (\emph{``Are models sensitive to the factors that modulate human pragmatic inferences?''}). We urge other NLP researchers to consider variability in human behaviors instead of relying on categorical generalizations \citep[see also][]{pavlick_inherent_2019,jiang_investigating_2022,baan_stop_2022,webson_are_2023}. Through this approach, we can build models that capture the rich variability of human language, and use these models to refine our theories about the human mind.

\section*{Acknowledgments}
We thank the anonymous reviewers as well as the action editor, Ehud Reiter, for their insightful feedback. J.H.~is supported by an NSF Graduate Research Fellowship (\#1745302) and an NSF Doctoral Dissertation Research Improvement Grant (BCS-2116918). S.S. is supported by the NSF under Grant \#2030859 to the
Computing Research Association for the CIFellows Project and the European Research Council (ERC) under the European Union's Horizon 2020 research and innovation programme (grant agreement No 948878). J.H.~and R.L.~also gratefully acknowledge support from the Simons Center for the Social Brain at MIT. Any opinions, findings, and conclusions or recommendations expressed in this material are those of the authors and do not necessarily reflect
the views of the National Science Foundation nor the Computing Research Association.

\bibliography{tacl2021}
\bibliographystyle{acl_natbib}

\iftaclpubformat

\onecolumn






  
\fi

\end{document}

%% file: figure1.tex
\begin{figure}[t]
    \centering
    \subfloat[\label{fig:example-within-scale}]{
        \includegraphics[width=0.96\linewidth]{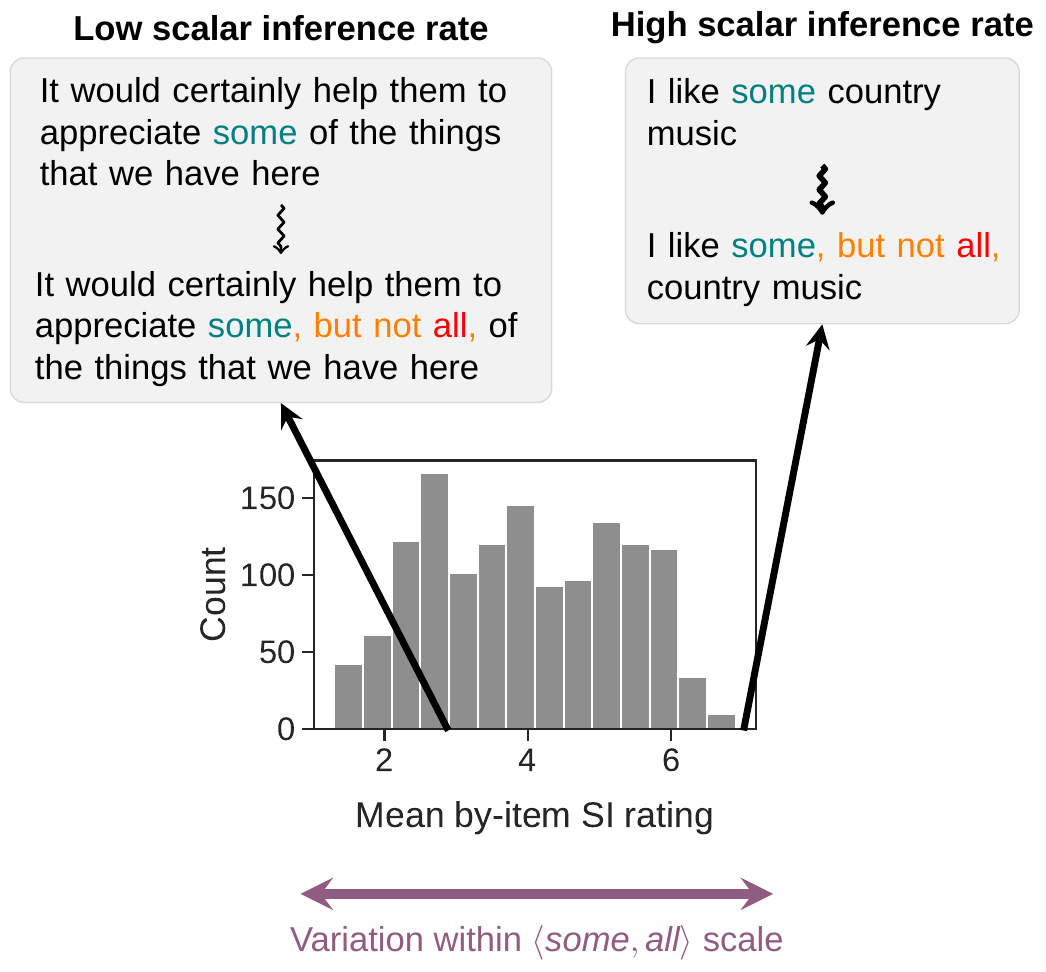}
    }\\
    \subfloat[\label{fig:example-cross-scale}]{
        \includegraphics[width=0.95\linewidth]{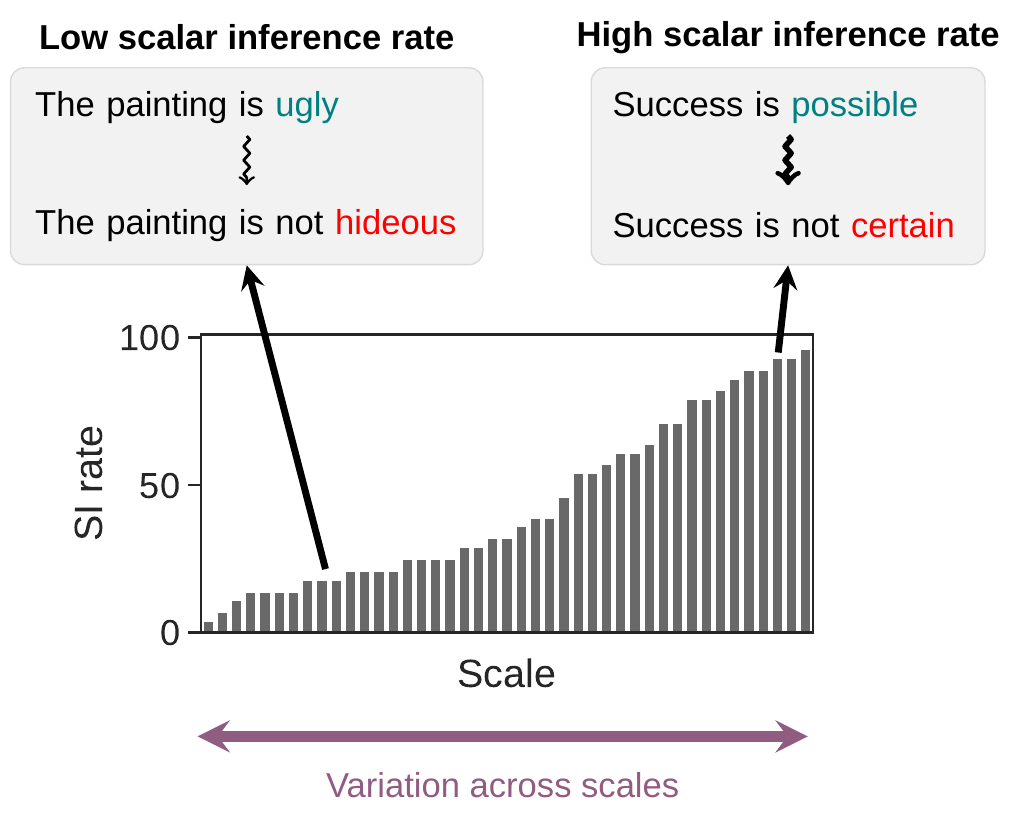}
    }
    \caption{(a) Distribution of human scalar inference (SI) ratings (on scale of 1-7) across instances of the \scale{some}{all} scale (reproduction of Fig.~1, \citealt{degen_investigating_2015}). (b) Average SI rates across scales formed by different lexical items (reproduction of Fig.~2, \citealt{van_tiel_scalar_2016}).}
    \label{fig:example}
\end{figure}

%% file: rx_dists_figure.tex
\begin{figure*}[t]
    \centering
    \begin{tikzpicture}
    \node (pic) {\includegraphics[width=\linewidth]{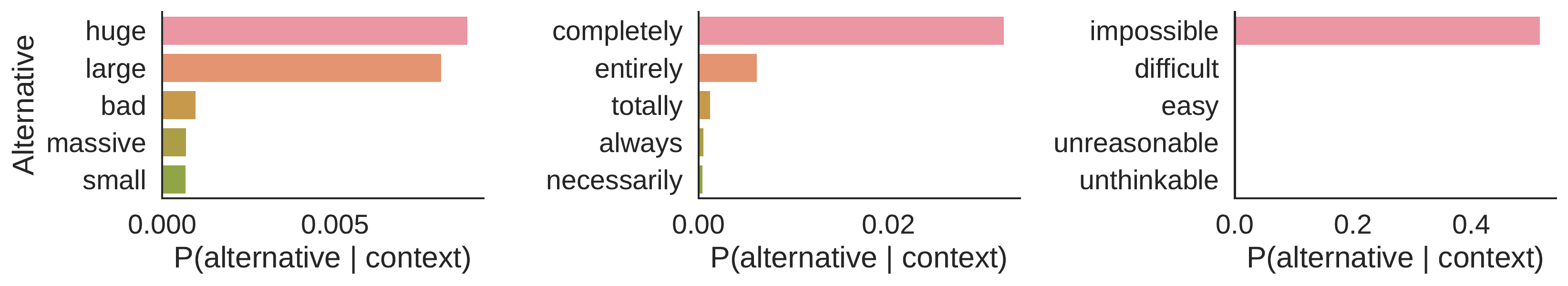}};
    \node[rounded corners,fill=black!5,align=left,font=\scriptsize] at (-4.8,1.9) {\texttt{The elephant is {\color{teal}{big}},}\\\texttt{but not {\color{red}{\underline{enormous}}}}};
    \node[rounded corners,fill=black!5,align=left,font=\scriptsize] at (0.2,1.9) {\texttt{The coast is {\color{teal}{largely}} flooded,}\\\texttt{but not {\color{red}{\underline{totally}}}}};
    \node[rounded corners,fill=black!5,align=left,font=\scriptsize] at (6.1,1.9) {\texttt{The problem is {\color{teal}{hard}},}\\\texttt{but not {\color{red}{\underline{unsolvable}}}}};
    \end{tikzpicture}
    \caption{Probability assigned by GPT-2 to top 5 candidate strong alternatives (y-axis) for 3 example weak scalar items: \scalemate{big}, \scalemate{largely}, and \scalemate{hard} \citep{ronai_three_2022}. The full scalar construction is shown above each subplot, with the original tested strong scalemate underlined in {\color{red}{\underline{red}}}.}
    \label{fig:rx-dists}
\end{figure*}
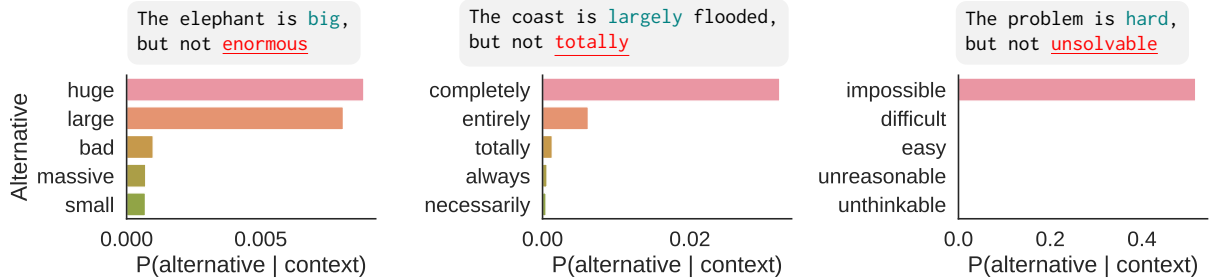

%% file: tacl2021v1-template_FOR_ARXIV.bbl
\begin{thebibliography}{67}
\expandafter\ifx\csname natexlab\endcsname\relax\def\natexlab#1{#1}\fi

\bibitem[{Baan et~al.(2022)Baan, Aziz, Plank, and Fernandez}]{baan_stop_2022}
Joris Baan, Wilker Aziz, Barbara Plank, and Raquel Fernandez. 2022.
\newblock \href {https://aclanthology.org/2022.emnlp-main.124} {Stop
  {Measuring} {Calibration} {When} {Humans} {Disagree}}.
\newblock In \emph{Proceedings of the 2022 {Conference} on {Empirical}
  {Methods} in {Natural} {Language} {Processing}}, pages 1892--1915, Abu Dhabi,
  United Arab Emirates. Association for Computational Linguistics.

\bibitem[{Beltrama and Xiang(2013)}]{beltrama_is_2013}
Andrea Beltrama and Ming Xiang. 2013.
\newblock \href
  {https://ojs.ub.uni-konstanz.de/sub/index.php/sub/article/view/333} {Is
  `good' better than `excellent'? {An} experimental investigation on scalar
  implicatures and gradable adjectives}.
\newblock \emph{Proceedings of Sinn und Bedeutung}, 17.

\bibitem[{Breheny et~al.(2018)Breheny, Klinedinst, Romoli, and
  Sudo}]{breheny_symmetry_2018}
Richard Breheny, Nathan Klinedinst, Jacopo Romoli, and Yasutada Sudo. 2018.
\newblock \href {https://doi.org/10.1007/s11050-017-9141-z} {The symmetry
  problem: current theories and prospects}.
\newblock \emph{Natural Language Semantics}, 26(2):85--110.

\bibitem[{Buccola et~al.(2021)Buccola, Križ, and
  Chemla}]{buccola_conceptual_2021}
Brian Buccola, Manuel Križ, and Emmanuel Chemla. 2021.
\newblock \href {https://doi.org/10.1007/s10988-021-09327-w} {Conceptual
  alternatives: {Competition} in language and beyond}.
\newblock \emph{Linguistics and Philosophy}.

\bibitem[{Bybee and Beckner(2015)}]{bybee_usage-based_2015}
Joan~L. Bybee and Clay Beckner. 2015.
\newblock Usage-based theory.
\newblock In Bernd Heine and Heiko Narrog, editors, \emph{The {Oxford}
  {Handbook} of {Linguistic} {Analysis}}. Oxford University Press.

\bibitem[{Degen(2015)}]{degen_investigating_2015}
Judith Degen. 2015.
\newblock \href {https://doi.org/10.3765/sp.8.11} {Investigating the
  distribution of \emph{some} (but not \emph{all}) implicatures using corpora
  and web-based methods}.
\newblock \emph{Semantics and Pragmatics}, 8(11):1--55.

\bibitem[{Degen and Tanenhaus(2015)}]{degen_processing_2015}
Judith Degen and Michael~K. Tanenhaus. 2015.
\newblock \href {https://doi.org/10.1111/cogs.12171} {Processing {Scalar}
  {Implicature}: {A} {Constraint}-{Based} {Approach}}.
\newblock \emph{Cognitive Science}, 39(4):667--710.

\bibitem[{Degen and Tanenhaus(2016)}]{degen_availability_2016}
Judith Degen and Michael~K. Tanenhaus. 2016.
\newblock \href {https://doi.org/10.1111/cogs.12227} {Availability of
  {Alternatives} and the {Processing} of {Scalar} {Implicatures}: {A} {Visual}
  {World} {Eye}-{Tracking} {Study}}.
\newblock \emph{Cognitive Science}, 40(1):172--201.
\newblock Publisher: John Wiley \& Sons, Ltd.

\bibitem[{Degen et~al.(2015)Degen, Tessler, and Goodman}]{degen_wonky_2015}
Judith Degen, Michael~Henry Tessler, and Noah~D. Goodman. 2015.
\newblock \href {https://cocolab.stanford.edu/papers/DegenEtAl2015-Cogsci.pdf}
  {Wonky worlds: {Listeners} revise world knowledge when utterances are odd}.
\newblock In \emph{Proceedings of the 37th {Annual} {Meeting} of the
  {Cognitive} {Science} {Society}}.

\bibitem[{Devlin et~al.(2019)Devlin, Chang, Lee, and
  Toutanova}]{devlin_bert_2019}
Jacob Devlin, Ming-Wei Chang, Kenton Lee, and Kristina Toutanova. 2019.
\newblock \href {https://doi.org/10.18653/v1/N19-1423} {{BERT}: {Pre}-training
  of {Deep} {Bidirectional} {Transformers} for {Language} {Understanding}}.
\newblock In \emph{Proceedings of the 2019 {Conference} of the {North}
  {American} {Chapter} of the {Association} for {Computational} {Linguistics}:
  {Human} {Language} {Technologies}, {Volume} 1 ({Long} and {Short} {Papers})},
  pages 4171--4186, Minneapolis, Minnesota. Association for Computational
  Linguistics.

\bibitem[{Doran et~al.(2009)Doran, Baker, McNabb, Larson, and
  Ward}]{doran_non-unified_2009}
Ryan Doran, Rachel~E. Baker, Yaron McNabb, Meredith Larson, and Gregory Ward.
  2009.
\newblock \href
  {https://doi.org/https://doi.org/10.1163/187730909X12538045489854} {On the
  {Non}-{Unified} {Nature} of {Scalar} {Implicature}: {An} {Empirical}
  {Investigation}}.
\newblock \emph{International Review of Pragmatics}, 1(2):211 -- 248.
\newblock Place: Leiden, The Netherlands Publisher: Brill.

\bibitem[{Eiteljoerge et~al.(2018)Eiteljoerge, Pouscoulous, and
  Lieven}]{eiteljoerge_pieces_2018}
Sarah F.~V. Eiteljoerge, Nausicaa Pouscoulous, and Elena V.~M. Lieven. 2018.
\newblock \href {https://doi.org/10.3389/fpsyg.2018.01928} {Some {Pieces} {Are}
  {Missing}: {Implicature} {Production} in {Children}}.
\newblock \emph{Frontiers in Psychology}, 9:1928.

\bibitem[{Fox and Katzir(2011)}]{fox_characterization_2011}
Danny Fox and Roni Katzir. 2011.
\newblock \href {https://doi.org/10.1007/s11050-010-9065-3} {On the
  characterization of alternatives}.
\newblock \emph{Natural Language Semantics}, 19(1):87--107.
\newblock ISBN: 1572-865X.

\bibitem[{Frank and Goodman(2012)}]{frank_predicting_2012}
Michael~C. Frank and Noah~D. Goodman. 2012.
\newblock \href {https://doi.org/10.1126/science.1218633} {Predicting
  {Pragmatic} {Reasoning} in {Language} {Games}}.
\newblock \emph{Science}, 336(6084):998--998.

\bibitem[{Frank and Bod(2011)}]{frank_insensitivity_2011}
Stefan~L. Frank and Rens Bod. 2011.
\newblock \href {https://doi.org/10.1177/0956797611409589} {Insensitivity of
  the {Human} {Sentence}-{Processing} {System} to {Hierarchical} {Structure}}.
\newblock \emph{Psychological Science}, 22(6):829--834.

\bibitem[{Futrell et~al.(2019)Futrell, Wilcox, Morita, Qian, Ballesteros, and
  Levy}]{futrell_neural_2019}
Richard Futrell, Ethan Wilcox, Takashi Morita, Peng Qian, Miguel Ballesteros,
  and Roger Levy. 2019.
\newblock \href {https://doi.org/10.18653/v1/N19-1004} {Neural language models
  as psycholinguistic subjects: {Representations} of syntactic state}.
\newblock In \emph{Proceedings of the 2019 {Conference} of the {North}
  {American} {Chapter} of the {Association} for {Computational} {Linguistics}:
  {Human} {Language} {Technologies}, {Volume} 1 ({Long} and {Short} {Papers})},
  pages 32--42, Minneapolis, Minnesota. Association for Computational
  Linguistics.

\bibitem[{Garí~Soler and Apidianaki(2020)}]{gari_soler_bert_2020}
Aina Garí~Soler and Marianna Apidianaki. 2020.
\newblock \href {https://doi.org/10.18653/v1/2020.emnlp-main.598} {{BERT}
  {Knows} {Punta} {Cana} is not just beautiful, it's gorgeous: {Ranking}
  {Scalar} {Adjectives} with {Contextualised} {Representations}}.
\newblock In \emph{Proceedings of the 2020 {Conference} on {Empirical}
  {Methods} in {Natural} {Language} {Processing} ({EMNLP})}, pages 7371--7385,
  Online. Association for Computational Linguistics.

\bibitem[{Garí~Soler and Apidianaki(2021)}]{gari_soler_scalar_2021}
Aina Garí~Soler and Marianna Apidianaki. 2021.
\newblock \href {https://doi.org/10.18653/v1/2021.naacl-main.370} {Scalar
  {Adjective} {Identification} and {Multilingual} {Ranking}}.
\newblock In \emph{Proceedings of the 2021 {Conference} of the {North}
  {American} {Chapter} of the {Association} for {Computational} {Linguistics}:
  {Human} {Language} {Technologies}}, pages 4653--4660, Online. Association for
  Computational Linguistics.

\bibitem[{Gauthier et~al.(2020)Gauthier, Hu, Wilcox, Qian, and
  Levy}]{gauthier-etal-2020-syntaxgym}
Jon Gauthier, Jennifer Hu, Ethan Wilcox, Peng Qian, and Roger Levy. 2020.
\newblock \href {https://doi.org/10.18653/v1/2020.acl-demos.10} {{S}yntax{G}ym:
  An online platform for targeted evaluation of language models}.
\newblock In \emph{Proceedings of the 58th Annual Meeting of the Association
  for Computational Linguistics: System Demonstrations}, pages 70--76, Online.
  Association for Computational Linguistics.

\bibitem[{Gazdar(1979)}]{gazdar_pragmatics_1979}
Gerald Gazdar. 1979.
\newblock \emph{Pragmatics: {Implicature}, presupposition, and logical form}.
\newblock Academic Press, New York.

\bibitem[{Godfrey et~al.(1992)Godfrey, Holliman, and
  McDaniel}]{godfrey_switchboard_1992}
J.J. Godfrey, E.C. Holliman, and J.~McDaniel. 1992.
\newblock Switchboard: {A} telephone speech corpus for research and
  development.
\newblock In \emph{International {Conferenceon} {Acoustics}, {Speech} and
  {Signal} {Processing}}, pages 517--520.

\bibitem[{Goodman and Frank(2016)}]{goodman_pragmatic_2016}
Noah~D. Goodman and Michael~C. Frank. 2016.
\newblock \href {https://doi.org/10.1016/j.tics.2016.08.005} {Pragmatic
  {Language} {Interpretation} as {Probabilistic} {Inference}}.
\newblock \emph{Trends in Cognitive Sciences}, 20(11):818--829.

\bibitem[{Goodman and Lassiter(2015)}]{goodman_probabilistic_2015}
Noah~D. Goodman and Daniel Lassiter. 2015.
\newblock \href {https://doi.org/10.1002/9781118882139.ch21} {Probabilistic
  {Semantics} and {Pragmatics}: {Uncertainty} in {Language} and {Thought}}.
\newblock In \emph{The {Handbook} of {Contemporary} {Semantic} {Theory}}, pages
  655--686. John Wiley \& Sons, Ltd.

\bibitem[{Gotzner et~al.(2018)Gotzner, Solt, and Benz}]{gotzner_scalar_2018}
Nicole Gotzner, Stephanie Solt, and Anton Benz. 2018.
\newblock \href {https://doi.org/10.3389/fpsyg.2018.01659} {Scalar {Diversity},
  {Negative} {Strengthening}, and {Adjectival} {Semantics}}.
\newblock \emph{Frontiers in Psychology}, 9:1659.

\bibitem[{Grice(1975)}]{grice_logic_1975}
Herbert~P. Grice. 1975.
\newblock \href {http://www.ucl.ac.uk/ls/studypacks/Grice-Logic.pdf} {Logic and
  {Conversation}}.
\newblock In Peter Cole and Jerry~L. Morgan, editors, \emph{Syntax and
  {Semantics}: {Speech} {Acts}}, volume~3, pages 41--58. Academic Press.

\bibitem[{Hearst(1992)}]{hearst_automatic_1992}
Marti~A. Hearst. 1992.
\newblock \href {https://aclanthology.org/C92-2082} {Automatic {Acquisition} of
  {Hyponyms} from {Large} {Text} {Corpora}}.
\newblock In \emph{{COLING} 1992 {Volume} 2: {The} 14th {International}
  {Conference} on {Computational} {Linguistics}}.

\bibitem[{Horn(1989)}]{horn_natural_1989}
Laurence~R. Horn. 1989.
\newblock \emph{A {Natural} {History} of {Negation}}.
\newblock Chicago University Press.

\bibitem[{Hu et~al.(2020)Hu, Gauthier, Qian, Wilcox, and
  Levy}]{hu_systematic_2020}
Jennifer Hu, Jon Gauthier, Peng Qian, Ethan Wilcox, and Roger Levy. 2020.
\newblock \href {https://doi.org/10.18653/v1/2020.acl-main.158} {A {Systematic}
  {Assessment} of {Syntactic} {Generalization} in {Neural} {Language}
  {Models}}.
\newblock In \emph{Proceedings of the 58th {Annual} {Meeting} of the
  {Association} for {Computational} {Linguistics}}, pages 1725--1744, Online.
  Association for Computational Linguistics.

\bibitem[{Jeretic et~al.(2020)Jeretic, Warstadt, Bhooshan, and
  Williams}]{jeretic_are_2020}
Paloma Jeretic, Alex Warstadt, Suvrat Bhooshan, and Adina Williams. 2020.
\newblock \href {https://doi.org/10.18653/v1/2020.acl-main.768} {Are {Natural}
  {Language} {Inference} {Models} {IMPPRESsive}? {Learning} {IMPlicature} and
  {PRESupposition}}.
\newblock In \emph{Proceedings of the 58th {Annual} {Meeting} of the
  {Association} for {Computational} {Linguistics}}, pages 8690--8705, Online.
  Association for Computational Linguistics.

\bibitem[{Jiang and Marneffe(2022)}]{jiang_investigating_2022}
Nan-Jiang Jiang and Marie-Catherine~de Marneffe. 2022.
\newblock \href {https://doi.org/10.1162/tacl_a_00523} {Investigating {Reasons}
  for {Disagreement} in {Natural} {Language} {Inference}}.
\newblock \emph{Transactions of the Association for Computational Linguistics},
  10:1357--1374.

\bibitem[{Katzir(2007)}]{katzir_structurally-defined_2007}
Roni Katzir. 2007.
\newblock \href {https://doi.org/10.1007/s10988-008-9029-y}
  {Structurally-defined alternatives}.
\newblock \emph{Linguistics and Philosophy}, 30(6):669--690.

\bibitem[{Kim and de~Marneffe(2013)}]{kim_deriving_2013}
Joo-Kyung Kim and Marie-Catherine de~Marneffe. 2013.
\newblock \href {https://aclanthology.org/D13-1169} {Deriving {Adjectival}
  {Scales} from {Continuous} {Space} {Word} {Representations}}.
\newblock In \emph{Proceedings of the 2013 {Conference} on {Empirical}
  {Methods} in {Natural} {Language} {Processing}}, pages 1625--1630, Seattle,
  Washington, USA. Association for Computational Linguistics.

\bibitem[{Kroch(1972)}]{kroch_lexical_1972}
Anthony Kroch. 1972.
\newblock Lexical and inferred meanings for some time adverbs.
\newblock \emph{Quarterly Progress Reports of the Research Laboratory of
  Electronics}, 104:260--267.

\bibitem[{Lassiter(2022)}]{lassiter_how_2022}
Daniel Lassiter. 2022.
\newblock How not to identify a scalar implicature ({The} importance of
  priors).
\newblock Presentation at Cognitive Semantic and Quantities Workshop,
  University of Amsterdam.

\bibitem[{Levinson(2000)}]{levinson_presumptive_2000}
Stephen Levinson. 2000.
\newblock \emph{Presumptive meaning: {The} theory of generalized conversational
  implicature}.
\newblock MIT Press.

\bibitem[{Li et~al.(2021)Li, Schuster, and Degen}]{li_predicting_2021}
Elissa Li, Sebastian Schuster, and Judith Degen. 2021.
\newblock \href {https://doi.org/10.7275/xr01-a852} {Predicting {Scalar}
  {Inferences} {From} "{Or}" to "{Not} {Both}" {Using} {Neural} {Sentence}
  {Encoders}}.
\newblock In \emph{Proceedings of the {Society} for {Computation} in
  {Linguistics}}, volume~4.

\bibitem[{Linzen et~al.(2016)Linzen, Dupoux, and
  Goldberg}]{linzen_assessing_2016}
Tal Linzen, Emmanuel Dupoux, and Yoav Goldberg. 2016.
\newblock \href {https://doi.org/10.1162/tacl_a_00115} {Assessing the {Ability}
  of {LSTMs} to {Learn} {Syntax}-{Sensitive} {Dependencies}}.
\newblock \emph{Transactions of the Association for Computational Linguistics},
  4:521--535.

\bibitem[{Lison and Tiedemann(2016)}]{lison_opensubtitles2016_2016}
Pierre Lison and Jörg Tiedemann. 2016.
\newblock \href
  {http://www.lrec-conf.org/proceedings/lrec2016/pdf/947_Paper.pdf}
  {{OpenSubtitles2016}: {Extracting} {Large} {Parallel} {Corpora} from {Movie}
  and {TV} {Subtitles}}.
\newblock In \emph{Proceedings of the 10th {International} {Conference} on
  {Language} {Resources} and {Evaluation}}.

\bibitem[{Loper and Bird(2002)}]{loper_nltk_2002}
Edward Loper and Steven Bird. 2002.
\newblock \href {https://doi.org/10.3115/1118108.1118117} {{NLTK}: {The}
  {Natural} {Language} {Toolkit}}.
\newblock In \emph{Proceedings of the {ACL}-02 {Workshop} on {Effective}
  {Tools} and {Methodologies} for {Teaching} {Natural} {Language} {Processing}
  and {Computational} {Linguistics}}, pages 63--70, Philadelphia, Pennsylvania,
  USA. Association for Computational Linguistics.

\bibitem[{de~Marneffe et~al.(2010)de~Marneffe, Manning, and
  Potts}]{de_marneffe_was_2010}
Marie-Catherine de~Marneffe, Christopher~D. Manning, and Christopher Potts.
  2010.
\newblock \href {https://www.aclweb.org/anthology/P10-1018} {“{Was} {It}
  {Good}? {It} {Was} {Provocative}.” {Learning} the {Meaning} of {Scalar}
  {Adjectives}}.
\newblock In \emph{Proceedings of the 48th {Annual} {Meeting} of the
  {Association} for {Computational} {Linguistics}}, pages 167--176, Uppsala,
  Sweden. Association for Computational Linguistics.

\bibitem[{Marr(1982)}]{marr_vision_1982}
David Marr. 1982.
\newblock \emph{Vision: {A} {Computational} {Approach}}.
\newblock Freeman \& Co., San Francisco.

\bibitem[{de~Melo and Bansal(2013)}]{de_melo_good_2013}
Gerard de~Melo and Mohit Bansal. 2013.
\newblock \href {https://doi.org/10.1162/tacl_a_00227} {Good, {Great},
  {Excellent}: {Global} {Inference} of {Semantic} {Intensities}}.
\newblock \emph{Transactions of the Association for Computational Linguistics},
  1:279--290.

\bibitem[{Merkx and Frank(2021)}]{merkx_human_2021}
Danny Merkx and Stefan~L. Frank. 2021.
\newblock \href {https://doi.org/10.18653/v1/2021.cmcl-1.2} {Human {Sentence}
  {Processing}: {Recurrence} or {Attention}?}
\newblock In \emph{Proceedings of the {Workshop} on {Cognitive} {Modeling} and
  {Computational} {Linguistics}}, pages 12--22, Online. Association for
  Computational Linguistics.

\bibitem[{Miller(1995)}]{miller_wordnet_1995}
G.A. Miller. 1995.
\newblock {WordNet}: {A} {Lexical} {Database} for {English}.
\newblock \emph{Communications of the ACM}, 38(11):39--41.

\bibitem[{van Miltenburg(2015)}]{van_miltenburg_detecting_2015}
Emiel van Miltenburg. 2015.
\newblock \href {https://arxiv.org/abs/1504.08102} {Detecting and ordering
  adjectival scalemates}.
\newblock In \emph{Proceedings of MAPLEX}, Yamagata, Japan.

\bibitem[{Pankratz and van Tiel(2021)}]{pankratz_role_2021}
Elizabeth Pankratz and Bob van Tiel. 2021.
\newblock \href {https://doi.org/10.1017/langcog.2021.13} {The role of
  relevance for scalar diversity: a usage-based approach}.
\newblock \emph{Language and Cognition}, 13(4):562--594.
\newblock Edition: 2021/08/16 Publisher: Cambridge University Press.

\bibitem[{Pavlick and Kwiatkowski(2019)}]{pavlick_inherent_2019}
Ellie Pavlick and Tom Kwiatkowski. 2019.
\newblock \href {https://doi.org/10.1162/tacl_a_00293} {Inherent
  {Disagreements} in {Human} {Textual} {Inferences}}.
\newblock \emph{Transactions of the Association for Computational Linguistics},
  7:677--694.

\bibitem[{Pennington et~al.(2014)Pennington, Socher, and
  Manning}]{pennington_glove_2014}
Jeffrey Pennington, Richard Socher, and Christopher Manning. 2014.
\newblock \href {https://doi.org/10.3115/v1/D14-1162} {{GloVe}: {Global}
  {Vectors} for {Word} {Representation}}.
\newblock In \emph{Proceedings of the 2014 {Conference} on {Empirical}
  {Methods} in {Natural} {Language} {Processing} ({EMNLP})}, pages 1532--1543,
  Doha, Qatar. Association for Computational Linguistics.

\bibitem[{Qing et~al.(2016)Qing, Goodman, and Lassiter}]{qing_rational_2016}
Ciyang Qing, Noah~D. Goodman, and Daniel Lassiter. 2016.
\newblock \href
  {http://cocolab.stanford.edu/papers/QingGoodmanLassiter2016-Cogsci.pdf} {A
  {Rational} {Speech}-{Act} {Model} of {Projective} {Content}}.
\newblock In \emph{Proceedings of the 38th Annual Meeting of the Cognitive
  Science Society}.

\bibitem[{Radford et~al.(2019)Radford, Wu, Child, Luan, Amodei, and
  Sutskever}]{radford_language_2019}
Alec Radford, Jeff Wu, Rewon Child, David Luan, Dario Amodei, and Ilya
  Sutskever. 2019.
\newblock \href
  {https://d4mucfpksywv.cloudfront.net/better-language-models/language-models.pdf}
  {Language {Models} are {Unsupervised} {Multitask} {Learners}}.

\bibitem[{Roberts(2012)}]{roberts_information_2012}
Craige Roberts. 2012.
\newblock \href {https://doi.org/10.3765/sp.5.6} {Information structure in
  discourse: {Towards} an integrated formal theory of pragmatics}.
\newblock \emph{Semantics and Pragmatics}, 5(6):1--69.

\bibitem[{Ronai and Xiang(2021)}]{ronai_exploring_2021}
Eszter Ronai and Ming Xiang. 2021.
\newblock \href {https://doi.org/10.3765/plsa.v6i1.5001} {Exploring the
  connection between {Question} {Under} {Discussion} and scalar diversity}.
\newblock In \emph{Proceedings of the {Linguistic} {Society} of {America}},
  volume~6, pages 649--662.

\bibitem[{Ronai and Xiang(2022)}]{ronai_three_2022}
Eszter Ronai and Ming Xiang. 2022.
\newblock \href
  {https://cpb-us-w2.wpmucdn.com/voices.uchicago.edu/dist/c/1271/files/2022/02/RonaiXiang_SuB26_paper.pdf}
  {Three factors in explaining scalar diversity}.
\newblock In \emph{Proceedings of {Sinn} und {Bedeutung} 26}.

\bibitem[{Rooth(1985)}]{rooth_association_1985}
Mats~E. Rooth. 1985.
\newblock \href {https://scholarworks.umass.edu/dissertations/AAI8509599/}
  {\emph{Association with {Focus} ({Montague} {Grammar}, semantics, only,
  even)}}.
\newblock {PhD} {Thesis}, University of Massachusetts.

\bibitem[{Schuster et~al.(2020)Schuster, Chen, and
  Degen}]{schuster_harnessing_2020}
Sebastian Schuster, Yuxing Chen, and Judith Degen. 2020.
\newblock \href {https://doi.org/10.18653/v1/2020.acl-main.479} {Harnessing the
  linguistic signal to predict scalar inferences}.
\newblock In \emph{Proceedings of the 58th {Annual} {Meeting} of the
  {Association} for {Computational} {Linguistics}}, pages 5387--5403, Online.
  Association for Computational Linguistics.

\bibitem[{Shivade et~al.(2015)Shivade, de~Marneffe, Fosler-Lussier, and
  Lai}]{shivade_corpus-based_2015}
Chaitanya Shivade, Marie-Catherine de~Marneffe, Eric Fosler-Lussier, and
  Albert~M. Lai. 2015.
\newblock \href {https://doi.org/10.3115/v1/N15-1051} {Corpus-based discovery
  of semantic intensity scales}.
\newblock In \emph{Proceedings of the 2015 {Conference} of the {North}
  {American} {Chapter} of the {Association} for {Computational} {Linguistics}:
  {Human} {Language} {Technologies}}, pages 483--493, Denver, Colorado.
  Association for Computational Linguistics.

\bibitem[{Sikos et~al.(2021)Sikos, Venhuizen, Drenhaus, and
  Crocker}]{sikos_reevaluating_2021}
Les Sikos, Noortje~J. Venhuizen, Heiner Drenhaus, and Matthew~W. Crocker. 2021.
\newblock \href {https://doi.org/10.1371/journal.pone.0248388} {Reevaluating
  pragmatic reasoning in language games}.
\newblock \emph{PLOS ONE}, 16(3):e0248388.

\bibitem[{Smith and Levy(2013)}]{smith_effect_2013}
Nathaniel~J. Smith and Roger Levy. 2013.
\newblock \href
  {https://doi.org/https://doi.org/10.1016/j.cognition.2013.02.013} {The effect
  of word predictability on reading time is logarithmic}.
\newblock \emph{Cognition}, 128(3):302 -- 319.

\bibitem[{Sperber and Wilson(1986)}]{sperber_relevance_1986}
Dan Sperber and Deirdre Wilson. 1986.
\newblock \href
  {https://monoskop.org/images/e/e6/Sperber_Dan_Wilson_Deirdre_Relevance_Communica_and_Cognition_2nd_edition_1996.pdf}
  {\emph{Relevance: {Communication} and {Cognition}}}.
\newblock Wiley-Blackwell.

\bibitem[{Sun et~al.(2018)Sun, Tian, and Breheny}]{sun_link_2018}
Chao Sun, Ye~Tian, and Richard Breheny. 2018.
\newblock \href {https://doi.org/10.3389/fpsyg.2018.02092} {A {Link} {Between}
  {Local} {Enrichment} and {Scalar} {Diversity}}.
\newblock \emph{Frontiers in Psychology}, 9:2092.

\bibitem[{van Tiel et~al.(2016)van Tiel, van Miltenburg, Zevakhina, and
  Geurts}]{van_tiel_scalar_2016}
Bob van Tiel, Emiel van Miltenburg, Natalia Zevakhina, and Bart Geurts. 2016.
\newblock \href {https://doi.org/10.1093/jos/ffu017} {Scalar {Diversity}}.
\newblock \emph{Journal of Semantics}, 33(1):137--175.

\bibitem[{Tomasello(2003)}]{tomasello_constructing_2003}
Michael Tomasello. 2003.
\newblock \emph{Constructing a language: {A} usage-based theory of language
  acquisition}.
\newblock Harvard University Press.

\bibitem[{Webson et~al.(2023)Webson, Loo, Yu, and Pavlick}]{webson_are_2023}
Albert Webson, Alyssa~Marie Loo, Qinan Yu, and Ellie Pavlick. 2023.
\newblock \href {https://doi.org/10.48550/ARXIV.2301.07085} {Are {Language}
  {Models} {Worse} than {Humans} at {Following} {Prompts}? {It}'s
  {Complicated}}.
\newblock ArXiv preprint.

\bibitem[{Westera and Boleda(2020)}]{westera_closer_2020}
Matthijs Westera and Gemma Boleda. 2020.
\newblock \href {https://doi.org/10.18148/sub/2020.v24i2.908} {A closer look at
  scalar diversity using contextualized semantic similarity}.
\newblock \emph{Proceedings of Sinn und Bedeutung}, 24(2):439--454.

\bibitem[{Wilcox et~al.(2020)Wilcox, Gauthier, Hu, Qian, and
  Levy}]{wilcox_predictive_2020}
Ethan Wilcox, Jon Gauthier, Jennifer Hu, Peng Qian, and Roger Levy. 2020.
\newblock \href {https://cogsci.mindmodeling.org/2020/papers/0375/index.html}
  {On the predictive power of neural language models for human real-time
  comprehension behavior}.
\newblock In \emph{Proceedings of the 42nd Annual Meeting of the Cognitive
  Science Society}.

\bibitem[{Wolf et~al.(2020)Wolf, Debut, Sanh, Chaumond, Delangue, Moi, Cistac,
  Rault, Louf, Funtowicz, Davison, Shleifer, von Platen, Ma, Jernite, Plu, Xu,
  Le~Scao, Gugger, Drame, Lhoest, and Rush}]{wolf_transformers_2020}
Thomas Wolf, Lysandre Debut, Victor Sanh, Julien Chaumond, Clement Delangue,
  Anthony Moi, Pierric Cistac, Tim Rault, Remi Louf, Morgan Funtowicz, Joe
  Davison, Sam Shleifer, Patrick von Platen, Clara Ma, Yacine Jernite, Julien
  Plu, Canwen Xu, Teven Le~Scao, Sylvain Gugger, Mariama Drame, Quentin Lhoest,
  and Alexander Rush. 2020.
\newblock \href {https://doi.org/10.18653/v1/2020.emnlp-demos.6} {Transformers:
  {State}-of-the-{Art} {Natural} {Language} {Processing}}.
\newblock In \emph{Proceedings of the 2020 {Conference} on {Empirical}
  {Methods} in {Natural} {Language} {Processing}: {System} {Demonstrations}},
  pages 38--45, Online. Association for Computational Linguistics.

\bibitem[{Zhang et~al.(2023)Zhang, Bergen, Paunov, Ryskin, and
  Gibson}]{zhang2023scalar}
Zheng Zhang, Leon Bergen, Alexander Paunov, Rachel Ryskin, and Edward Gibson.
  2023.
\newblock \href {https://doi.org/https://doi.org/10.1111/cogs.13238} {Scalar
  implicature is sensitive to contextual alternatives}.
\newblock \emph{Cognitive Science}, 47(2).

\end{thebibliography}
